\documentclass[runningheads]{llncs}
\usepackage[T1]{fontenc}
\usepackage{graphicx}

\usepackage{todonotes}
\usepackage{amsmath,amssymb,amsfonts,xcolor,graphicx,wrapfig}
\usepackage{algorithmicx,algorithm}
\usepackage[noend]{algpseudocode}
\usepackage{mathtools}
\usepackage{booktabs}
\usepackage{adjustbox}
\usepackage{xspace}
\usepackage{color,colortbl}
\usepackage{multirow}
\usepackage{algorithm}
\usepackage{algpseudocode}
\usepackage[capitalise]{cleveref}
\usepackage[compatibility=false]{caption}
\usepackage{subcaption}
\usepackage{placeins} 

\usepackage{booktabs, multirow} 
\usepackage{siunitx}[=v2]
\usetikzlibrary{shapes,positioning,arrows,calc,automata,matrix,fit}
\tikzstyle{state}+=[minimum size = 6mm, inner sep=0,outer sep=1]
\tikzset{->,>=stealth'}
\input{format/macros}
\renewcommand{\paragraph}[1]{\smallskip\noindent{\bf#1}}

\begin{document}

\title{1--2--3--Go! Policy Synthesis for Parameterized Markov Decision Processes via Decision-Tree Learning and Generalization\thanks{
		This research was funded in part by the DFG project 427755713 GOPro, the DFG GRK 2428 (ConVeY), 
		the MUNI Award in Science and Humanities (MUNI/I/1757/2021) of the Grant Agency of Masaryk University, 
		and the EU under MSCA grant agreement 101034413 (IST-BRIDGE).}
}

\titlerunning{1--2--3--Go!}

\author{Muqsit Azeem\inst{1} \and 
		Debraj Chakraborty\inst{3} \and
		Sudeep Kanav\inst{3} \and
		Jan K\v{r}et\'{i}nsk\'{y}\inst{1,3} \and
		Mohammadsadegh Mohagheghi\inst{2} \and 
		Stefanie Mohr\inst{1} \and
		Maximilian Weininger\inst{4}}
\institute{Technical University of Munich, Munich, Germany \and
		Vali-e-Asr University of Rafsanjan, Rafsanjan, Iran \and
	 Masaryk University, Brno, Czech Republic \and
	 Institute of Science and Technology, Vienna, Austria}
\authorrunning{Azeem et al.}

\maketitle

\begin{abstract}
	Despite the advances in probabilistic model checking, the scalability of the verification methods remains limited.
	In particular, the state space often becomes extremely large when instantiating parameterized Markov decision processes (MDPs) even with moderate values.
	Synthesizing policies for such \emph{huge} MDPs is beyond the reach of available tools.
	We propose a learning-based approach to obtain a reasonable policy for such huge MDPs.
	
	The idea is to generalize optimal policies obtained by model-checking small instances to larger ones using decision-tree learning.
	Consequently, our method bypasses the need for explicit state-space exploration of large models, providing a practical solution to the state-space explosion problem.
	We demonstrate the efficacy of our approach by performing extensive experimentation on the
	relevant models from the quantitative verification benchmark set.
	The experimental results indicate that our policies perform well,
	even when the size of the model is orders of magnitude beyond the reach of state-of-the-art analysis tools.
	
\end{abstract}

\section{Introduction}
\paragraph{Markov decision processes (MDPs)} are \emph{the} model for combining probabilistic uncertainty and non-determinism. 
MDPs come with a rich theory and algorithmics developed over several decades with mature verification tools arising 20 years ago~\cite{Prism1} and proliferating since then~\cite{QComp}.
Despite all this effort, the \emph{scalability} of the methods is considerably worse than of those used for verification of non-deterministic systems with no probabilities, even for basic problems.

\paragraph{What to do about very large models?}
Researchers have made various attempts to tackle this issue, however, only with limited success.
Firstly, prominent techniques which work well in \emph{non-probabilistic} verification, such as symbolic techniques \cite{Prism1}, abstraction \cite{DBLP:conf/qest/KwiatkowskaNP06}, and symmetry reduction \cite{DBLP:conf/fmco/GroesserB05},
are harder to apply efficiently in the probabilistic setting.
Secondly, \emph{``engineering''} improvements, such as the use of external storage \cite{HH15} or parallelization, help by a significant, but principally very limited factor.
Thirdly, there is a \emph{relaxation of the guarantees} on the precision and/or certainty of the result, which we describe in detail below.

The result of the analysis is typically a number (called \emph{the value}), such as the expected reward or the probability to reach a given state, maximized (or minimized) over all resolutions of non-deterministic choices (called \emph{policies, strategies, schedulers, adversaries, or controllers} in different applications). 
It is generally accepted that the precise number is not needed and an approximation is sufficient in most settings.
Interestingly, until a few years ago \cite{DBLP:conf/rp/HaddadM14,atva14,DBLP:conf/cav/Baier0L0W17}, typically only the \emph{under-approximations} were computed for the fundamental (maximization) problems, with no reliable over-approximations (with dual issues for minimization).

It is worth noting that over-approximating is inherently harder since reasoning that the value cannot be greater than $x$ involves the claim that \emph{all} policies induce a lower value. 
In contrast to this universal quantification, the existential one is sufficient for under-approximating: upon providing \emph{a} policy, its value forms automatically a lower bound, which is typically easier to compute.
Consequently, many \emph{best-effort} approaches, such as reinforcement learning (RL)~\cite{SB98} and lightweight statistical model checking~\cite{DBLP:conf/tacas/BuddeDHS18} simply try to find a good policy while giving only empirically good chances to be close to the optimum.

This is sufficient in the setting of (i) \emph{policy synthesis}, where a ``good enough'' (close to optimum), but not necessarily optimal, controller is sought,
or (ii) \emph{bug hunting and falsification}, where finding significant counter-examples cheaply is desirable.
However, the cal quality of the results relies on certain assumptions of these methods:
RL results suffer when the rewards in the model are sparse (e.g.,\ in the case of reachability)
and lightweight statistical model checking suffers when near-optimal policies are not abundant.

To summarize, synthesizing \textbf{practically good policies} is sufficient in many settings and also the only way when the systems are too large.
Yet,  when the system is extremely large, the available techniques either run out of resources or yield policies that are far from optimum (and close to random).

Examining the structure of large MDPs in standard benchmark sets, e.g.~\cite{QVBS}, 
reveals that their huge sizes are typically not due to astronomically large human-written code, but rather because the MDPs are \emph{parameterized} (e.g., by the number of participants in a protocol) and then instantiated with large values.
Accordingly, this paper proposes a new approach to scalable policy synthesis for parameterized MDPs.
Namely, it produces \textbf{\emph{good} policies for arbitrarily large instantiations of parameterized MDPs 
in particular those beyond the reach of any state-of-the-art tools}.
We focus on probabilistic reachability (i)~for simplicity and (ii)~because it is a fundamental building block for many other problems.

\paragraph{Our main idea} is to generalize the decisions taken by the optimal policies for the smaller instances:
Instead of investigating a huge MDP,
we synthesize optimal policies for the given parameterized MDP by instantiating it with \emph{small} numbers.
We then \emph{generalize} this information and learn a policy 
that can be applied to any instantiation with an arbitrarily larger number.
It is important to note that 
we generalize the corresponding \emph{decisions} (i.e., the policy itself),
not the \emph{values} of the states across different parameterizations.
Indeed, while the numeric values can differ vastly, the optimal behaviour is often similar in all instantiations.
In order to capture this regularity, we thus need a \emph{symbolic} representation of a policy, which applies to \emph{all} instantiations.
{Decision trees} (DT) can provide such a representation.
Moreover, since they can represent policies explainably \cite{BrazdilCCFK15,dtcontrol2} and capture the essence of the decisions, not just a list of state-action pairs, they generalize well.

As an illustrative task for our ``generalization'', consider a buggy mutual exclusion protocol with a high number of participants.
While finding the bug with many participants may be hard, an exhaustive investigation of the case with two participants may reveal a scenario violating the exclusion.
A similar scenario can then also happen with many participants where the choices of the remaining participants may be irrelevant. 
Consequently, the key \emph{decisions in the policy} to find bug with two participant
may also be used for finding the bug with multiple participants.

\paragraph{Our contribution} can be summarized as follows:
\begin{itemize}
	\item We provide a \emph{simple} and elegant way of computing practically good policies for parameterized MDPs of \emph{any} size (as long as some instantiations exist that are small enough so that some technique can be applied), in particular also orders of magnitude beyond the reach of any other methods.
		The method is based on generalizing\footnote{The nature of our \emph{generalization-based} policy synthesis is also portrayed by our quipping title ``1--2--3--Go!'': Find out what works for cases 1, 2, and 3, then ``Go!'' and apply it for arbitrary large values of the parameters.} policies via their decision-tree representations.
	The method scales \emph{constantly} in the parameter instantiation since it applies available techniques to a fixed number of \textit{small base instantiations}, and the large instantiation is never explicitly considered. 
	\item We demonstrate the efficacy of the method experimentally on standard benchmarks. 
	In particular, from the practical perspective, we observe that	our policies mostly achieve values that are \emph{close to the actual optimum}.
Note that, this in principle cannot be guaranteed for instantiations too large for precise methods to apply, which are exactly of our interest.
	Nevertheless, the often consistent results on the smaller instantiations convincingly substantiate the expectation that the policies perform well also for the large instantiations.
	\item
	Finally, comparing to the benchmarks where our policies do not perform so well,
	we identify aspects of the models indicating where our heuristic generalizes well
	and where either more tailored or completely different techniques are required.
\end{itemize}

It should be emphasized that we regard the simplicity of our approach rather as an advantage, making it easy to exploit.
While there is a body of work on policy representation (via post-processing them), the use of DT to compute policies is very limited (as described in the Related Work below) and, to the best of our knowledge, non-existent for computing/generalizing them for arbitrarily large systems. 
Altogether, this simple, yet efficient idea deserves to be finally explored.

\subsection*{Related work}

\paragraph{Symbolic approaches} are widely used as for alleviating the challenges of the state explosion problem~\cite{BK08}.
These approaches are based on data structures storing the information of a model compactly.
In particular, the multi-terminal version of BDDs (MTBDDs) has been developed for probabilistic model checking~\cite{KleinBCDDKMM18,maisonneuve2009automatic,Parker03}.
In a sense, our approach is also symbolic, since we represent the policy using a decision tree.
This data structure is most suitable for the goal of explainability, as argued in, e.g.,~\cite{dtcontrol2}.

\paragraph{Reduction techniques} try to reduce the state space of the model while the smaller model satisfies the same set of properties.
A symmetry reduction technique for probabilistic models has been proposed in~\cite{KwiatkowskaNP06} for systems with several symmetric components.
Probabilistic bisimulation is available for MDPs and discrete-time Markov chains (DTMCs) that reduce the original model to the smallest one that satisfies the same set of temporal logic formulae~\cite{GrooteVV18}.
Considering a subset of temporal logic formulae, more efficient techniques have been proposed in~\cite{Kamaleson18} for reducing the model to a smaller one.
Applying reduction on a high-level description before constructing the resulting model is available in~\cite{SmolkaKKFHK019,LomuscioP19}.

Further techniques improving scalability of traditional algorithms include the following.
Using \emph{secondary storage} in an efficient way to keep a sparse representation of a large model has been studied in~\cite{HH15}. 
\emph{Compositional techniques} have been developed for the verification of stochastic systems~\cite{Feng14,li2019compositional}. 
\emph{Prioritizing computation} can reduce running times in many case studies by using topological state ordering~\cite{KwiatkowskaPQ11,CiesinskiBGK08} or learnt prioritizing~\cite{atva14,MohagheghiS20,DBLP:journals/lmcs/KretinskyM20}.

All the above techniques help solving larger models, however, only up to a certain limit.
In contrast, our approach synthesizes policy that can be applied to arbitrarily large instances.

\paragraph{Statistical model checking (SMC)} is an alternative solution for approximating the quantitative properties~\cite{DBLP:conf/qest/HenriquesMZPC12,atva14,HartmannsT15} by running a set of simulations on the model to approximate the requested values,
while providing a confidence interval for the precision of computed values for discrete and continuous-time Markov chains (DTMCs and CTMCs). This is scalable since the number of samples does not depend on the size of the model. Still, the length of the simulations does.
Using SMC for MDPs faces the difficulty of resolving non-determinism. A smart-sampling method has been proposed in~\cite{DArgenioLST15} that considers a set of random policies, with some of them hopefully approximating the optimal one; however, this method cannot generally provide a confidence interval for the precision of computations~\cite{HartmannsT15}.

\paragraph{Machine learning} within formal verification of MDP has been widely studied for a decade since the seminal~\cite{DBLP:conf/qest/HenriquesMZPC12}. 

An $L^*$ learning approach has been developed in~\cite{TapplerA0EL19} to learn an MDP model efficiently. Neural networks and regression can be used to resolve non-determinism of large MDPs and provide the opportunity of applying SMC for this class of models~\cite{RatajW18,GrosH0KS20}.

\paragraph{Reinforcement learning} has also been adapted to the setting of verification with objectives such as reachability \cite{atva14,DBLP:conf/tacas/HahnPSSTW19},
but the sparsity of the rewards is still an issue affecting the scalability.
Still, prioritizing the subset of the states that has the biggest impact on the value can allow for verifying huge models if such a subset is small and easy to find~\cite{atva14,DBLP:journals/lmcs/KretinskyM20}.
Unlike reinforcement learning (RL), which suggests policies for specific models through random exploration, our approach generalizes policies for various instances by computing optimal policies on smaller models and generalizing them.
Pyeatt and Howe \cite{PH99} propose using decision trees to approximate value function for discounted rewards in reinforcement learning.
However, we learn a DT that is a valid policy for any instantiations of the parameterized MDP,
whereas the DT learned in \cite{PH99} is applicable only to the model under consideration.

\paragraph{Decision trees} have been used as a data structure for representing MDP policies ~\cite{dtcontrol2,BrazdilCCFK15}. 
Interestingly, while binary decision diagrams (BDD) may appear to the verification community as a suitable candidate,
it has been shown that DT are more appropriate if adequately used \cite{BrazdilCCFK15,dtcontrol2,MohagheghiS20} due to their ability to handle various predicates and complex relationships, enhancing explainability.
Another advantage of DTs is the ability to declare some inputs as uninteresting (``don’t care'' inputs), saving on size via semantics of the controller.

\section{Preliminaries}

We provide basic definitions in Section~\ref{sec:prelim-mdp}, then describe what it means for models to be parameterized and scalable in Section~\ref{sec:prelim-structure} and finally recall how decision trees can be used for representing policies in Section~\ref{sec:prelim-dt}.

\subsection{Markov decision processes with a reachability objective}\label{sec:prelim-mdp}

A \emph{probability distribution} over a discrete set $X$ is a function $\mu : X\rightarrow [0,1]$ where $\sum_{x\in X}\mu(X) = 1$. 
We denote the set of all probability distributions over $X$ by $\Distributions(X)$.

\begin{definition}
A (finite) \emph{Markov Decision Process} (MDP) is a tuple $\MDP = (\states, \actions, \trans, \initstate, \gStates)$ where 
$\states$ is a finite set of states, $\actions$ is a finite set of actions, overloading $\actions(s)$ to denote the (non-empty) set of enabled actions for every state $s \in \states$,
$\trans:\states\times \actions \rightarrow \Distributions(\states)$ is a probabilistic transition function mapping each state $s\in\states$ and enabled action $a \in \actions(s)$ to a probability distribution over successor states,
$\initstate \in \states$ is the initial state, and $\gStates \subseteq \states$ is the set of goal states.	
\end{definition}

A \emph{Markov chain} (MC) can be seen as an MDP where $\abs{\actions(s)}=1$ for all $s\in\states$, i.e.\ a system exhibiting only probabilistic behavior, but no non-determinism.

The \textbf{semantics} of an MDP are defined in the usual way by means of policies and paths in the induced Markov chain.
An infinite \emph{path} $\apath = s_1 s_2 \ldots \in \states^\omega$ is an infinite sequence of states. 
A \emph{policy} is a function $\policy~\colon~\states \to \Distributions(\actions)$ that, intuitively, prescribes for every state which action to play. 
The policy is called deterministic if in every state it selects a single action surely, otherwise it is randomized.
Note that we limit our definition of policies w.l.o.g.\ to those that are memoryless (history-independent), i.e.\ those that depend only on the current state, not on a whole path.
We denote the $i$-th state on a path as $\apath_i$, the set of all paths as $\paths$, and the set of all policies as $\policies$.
By using a policy $\policy$ to resolve all nondeterministic choices in an MDP $\MDP$, we obtain a Markov chain $\MDP^{\policy}$~\cite[Definition 10.92]{BK08}.
 This Markov chain induces a unique probability measure $\prob^\policy_s$ over paths starting in state $s$~\cite[Definition 10.10]{BK08}.

The \textbf{reachability objective} is included in our definition of MDP in the form of the initial state $\initstate$ and the set of goal states $\gStates$.
Intuitively, the \emph{value} $\val$ of a reachability objective is the optimal probability to reach some goal state when starting in the initial state; formally
$
\val(\MDP) = \opt_{\policy \in \policies} \prob^\policy_{\initstate} [\lozenge \gStates],
$
where $\opt \in \{\max,\min\}$ indicates whether we are trying to reach or avoid the set of goal states and $\lozenge \gStates = \{\apath \in \paths \mid \exists i. \apath_i \in \gStates\}$ denotes the set of all paths that reach a goal state. 
One can restrict this optimum to the deterministic memoryless
policies~\cite[Proposition 6.2.1]{DBLP:books/wi/Puterman94}.

\subsection{State space structure and scalable models}\label{sec:prelim-structure}

For learning (e.g.\ of DTs) to be effective, it is important that the state space of MDP is \emph{structured}, i.e.\ every state is a tuple of values of state variables. 
In other words, the state space of the system is not monolithic (e.g.\ states defined by a simple numbering), but in fact, there are multiple factors defining it, for example, time or protocol state. Each of these factors is represented by a state variable $v_i$ with domain $\domain_i$.
Thus, every state $s \in \states$ is in fact a tuple $(v_1,v_2,\ldots,v_n)$, where each $v_i \in \domain_i$ is the value of a state~variable.

A \emph{parameterized MDP} can be described as a variant of standard MDP where certain parameters are not fixed constants but instead can take different values within a \emph{parameter space}.
 These parameters can be associated to the state-space of the system (for example, lower or upper bound of a state variable)
 or transition dynamics (the probabilities can be functions of the parameter). 
 We provide the formal definition below.

\begin{definition}
	A \emph{parameterized MDP} is a tuple $M = (\states_\theta, \actions_\theta, \trans_\theta, \initstate_\theta, \gStates_\theta, \Theta)$, where:
	\begin{itemize}
		\item $\Theta$ is the parameter space,
		\item For each $\theta \in \Theta$, the tuple $(\states_\theta, \actions_\theta, \trans_\theta, \initstate_\theta, \gStates_\theta)$ defines an MDP instance, where:
		\begin{itemize}
			\item $\states_\theta$ is the set of states,
			\item $\actions_\theta$ is the set of actions,
			\item $\trans_\theta: \states_\theta \times \actions_\theta \rightarrow \Distributions(\states_\theta)$ is the probabilistic transition function,
			\item $\initstate_\theta \in \states_\theta$ is the initial state,
			\item $\gStates_\theta \subseteq \states_\theta$ is the set of goal states.
		\end{itemize}
	\end{itemize}
	In this framework, different parameter values $\theta \in \Theta$ yield different instances of the MDP, and the parameterization can affect the state space, transition dynamics, or both.
\end{definition}
 Intuitively, a parameterized MDP can be seen as a family of MDPs where different value of parameter gives different \emph{instance} of the MDP. 
 In particular, this typically makes the models \emph{scalable}: by increasing the values of parameters in the model description, one can scale up the size of the state space of the model.
 MDPs in the PRISM benchmark suite~\cite{PRISMben} and the quantitative verification benchmark set~\cite{QVBS} have these properties of being structured~and~scalable.
 
\begin{figure}[t]
 	\centering
 	\includegraphics[scale=.20]{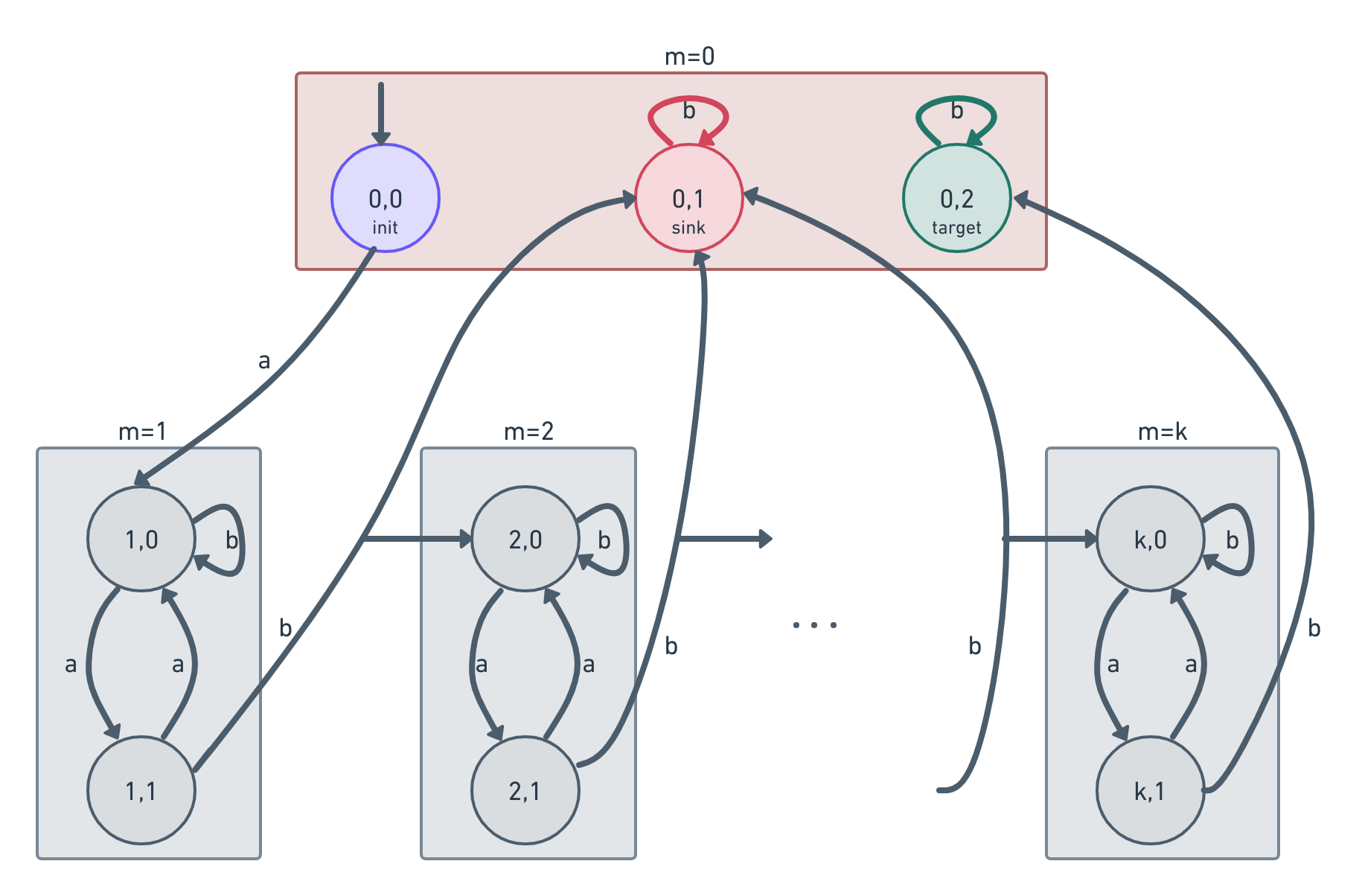}
 	\caption{A parameterized, scalable MDP with $k+1$ blocks, described in Example~\ref{ex:modules-mdp}.}
 	\label{fig:modules-mdp}
 \end{figure}

 \begin{example}\label{ex:modules-mdp}
 	Consider the MDP in Figure~\ref{fig:modules-mdp}. Every state is a tuple $(m,x)$ of two state variables $m$ and $x$ with domains $\domain_m = \{0,1,\ldots,k\}$ and $\domain_x = \{0,1,2\}$.
 	The state variable $m$ indicates which of the $k+1$ blocks we are in, while the state variable $x$ indicates the position inside a block.	
 	
 	The block with $m=0$ is special: it contains the initial state (0,0), the goal state (0,2), and a sink state (0,1) which cannot reach the goal. 
 	All other blocks look as follows: for every $m\in[1,k]$, the $x=0$ state can choose to continue to $x=1$ (action $a$) or self-loop (action $b$).
 	For every $m\in[1,k-1]$, the $x=1$ state can go back to $x=0$ in the same block (action $a$) or leave the block (action $b$). When using $b$, there is a 50\% chance of going to the sink state $(0,1)$ and a 50\% chance to continue to $x=0$ in the $(m+1)$-th block.
 	In the $k$-th block, the action leaving the block progresses to the goal state.
 	
 	The model is scalable, since the number of blocks $k$ can be increased arbitrarily.
 	This affects both the size of the state space, which is $2k+3$, as well as the maximum reachability probability, which is $0.5^{k-1}$. 
 \end{example}
 We assume the MDPs are defined using high-level modeling languages such as Probmela~\cite{BK08}, PRISM~\cite{PRISMben} or MODEST~\cite{DBLP:conf/fdl/Hartmanns12}. 
 In such modeling languages, MDPs are often represented as a composition  of multiple identical components, called \emph{modules}.
 For example, in a distributed system where multiple processes are interacting or sharing common resources, each process can be described as a separate module.
 In the context of this paper, we also consider the number of modules as a parameter. 

\begin{example}[Dining philosophers]\label{ex:dining-philosophers}
The dining philosophers problem involves a number of philosophers, seated around a circular table,
blessed with infinite availability of food and things to ponder about.
There is a fork placed between each pair of neighbouring philosophers and a philosopher must have both forks in order to eat. 
They may only pick up one fork at a time and once they finish eating, they place the forks back at the table and return to thinking. 

This can be described as an MDP where each philosopher is modeled as a separate module. Thus, the number of philosophers is a parameter.
\end{example}

\subsection{Decision trees for policy representation}\label{sec:prelim-dt}

Knowing that the state space is a product of state-variables, a deterministic policy is a mapping $\prod_i\domain_i\to\actions$ from tuples of state variables to actions.
By viewing the state variables as features and the actions as labels, we can employ machine learning classification techniques such as decision trees
, see e.g.~\cite[Chapter 3]{mitchell1997machine}, 
to represent a policy concisely.
We refer to~\cite{dtcontrol2} for an extensive description of the approach and its advantages.
Here, we shortly recall the most relevant definitions in order to formally state our results.

\begin{definition}
	A \emph{decision tree (DT)} $T$ is defined as follows:
	\begin{itemize}
		\item $T$ is a rooted full binary tree, meaning every node either is an \emph{inner node} and has exactly two children or is a \emph{leaf node} and has no children.
		\item Every inner node $v$ is associated with a decision predicate $\alpha_v$ which is a boolean function $\states \rightarrow \{\mathit{false}, \mathit{true}\}$ (or equivalently $\prod_i\domain_i\to \{\mathit{false}, \mathit{true}\}$).
		\item Every leaf $\ell$ is associated with an action $a_\ell \in A$.
	\end{itemize}
\end{definition}

For a given state $s$, we use the following recursive procedure to obtain the action $\policy(s)=a$ that a DT prescribes:
Start at the root. 
At an inner node, evaluate the decision predicate on the given state $s$. Depending on whether it evaluates to false or true, recursively continue evaluating on the left or right child, respectively.
At a leaf node, return the associated action $a$.

\begin{example}\label{ex:strat-explanation}
	Consider again the MDP given in Figure~\ref{fig:modules-mdp}.
	The optimal policy needs to continue towards the goal and not be stuck in any loops.
	This can be achieved by playing action $a$ in states where $x=0$ and action $b$ in all other states.
	Traditionally, this policy would be represented as a lookup table, storing $2k+3$ state-action pairs explicitly.
	Instead, we can condense the policy to the DT given in Figure~\ref{fig:dt-for-one-module-mdp}, mimicking the intuitive description of the policy: 
	If $x>0$, we play $b$, otherwise, we play $a$.
\end{example}

Constructing an optimal
binary decision tree is an NP-complete problem~\cite{HYAFIL197615}.
Consequently, practical decision tree learning algorithms are based on heuristics. But they tend to work reasonably well. 
Here, we briefly recall a general framework of learning the decision tree representation of a policy $\sigma$ 
as described in \cite{dtcontrol2}. 

If the policy suggests same action $a$ for all states (i.e., for all states $s$, we have $\sigma(s)=a$), the tree is just a single leaf node with label $a$.
Otherwise, we split the policy. 
A predicate $\rho$, defined on state variables, is chosen, and an inner node labeled $\rho$ is created. 
Then we partition the policy by evaluating the predicate on the state space, 
and recursively construct one
DT for the policy restricted to the states $\{s \in S | \rho(s)\}$
 where the predicate is
true, and one for the policy restricted to the states $\{s \in S | \neg\rho(s)\}$ where $\rho$ is false. 
These policies become
the children of the inner node with label $\rho$ and the process repeats recursively.

The selection of ``best'' predicate is done by selecting the one which is able to split the policy as homogeneous as possible. 
This is determined by optimizing some impurity measure such as Gini impurity~\cite[Chapter 4]{breiman1984classification} or entropy~\cite{shannon1948mathematical}. 

As we want the learnt DT to exactly represent the policy,
unlike other ML algorithms, 
we do not stop the learning early based on a stopping criterion.
Instead, we overfit on the data. 
So, the iteration stops when every state in the leaf node of the tree has the same labeling.

\section{Generalizing policies from small problem instances}\label{sec:approach}
In this section, we develop an approach for obtaining good policies for MDPs 
that are practically beyond the reach of \emph{any available} rigorous analysis.

Our approach exploits the regularity in structure of the MDPs,
therefore, we focus on parameterized MDPs where we expect regularity in the state space.
Intuitively, we solve a few small instances (colloquially speaking ``1, 2, and 3'')
where an optimal policy is easy to compute.
Then we generalize these policies by learning a DT from the combined information.
The policy represented by this DT is applicable to any instantiation of the parameterized MDP,
even the ones that are infeasibly large for any state-of-the-art solver.

More concretely, our approach proceeds in three phases, each of which is described in detail in the following subsections.
\begin{enumerate}
	\item Select the parameters for the small instances (\emph{base} instances) to learn on.
	\item Collect the optimal policies on the base instances.
	\item Generalize these policies by learning a DT.
\end{enumerate}
Finally, we discuss how to apply the DT to the MDP and evaluate the policy in order to judge its performance.
We illustrate every phase on our running example, the MDP in Figure~\ref{fig:modules-mdp}.

\subsection{Parameter selection}\label{sec:3params}
\begin{figure}[t]
	\centering
	\begin{subfigure}{0.66\textwidth}
		\centering
		\includegraphics[width=\textwidth]{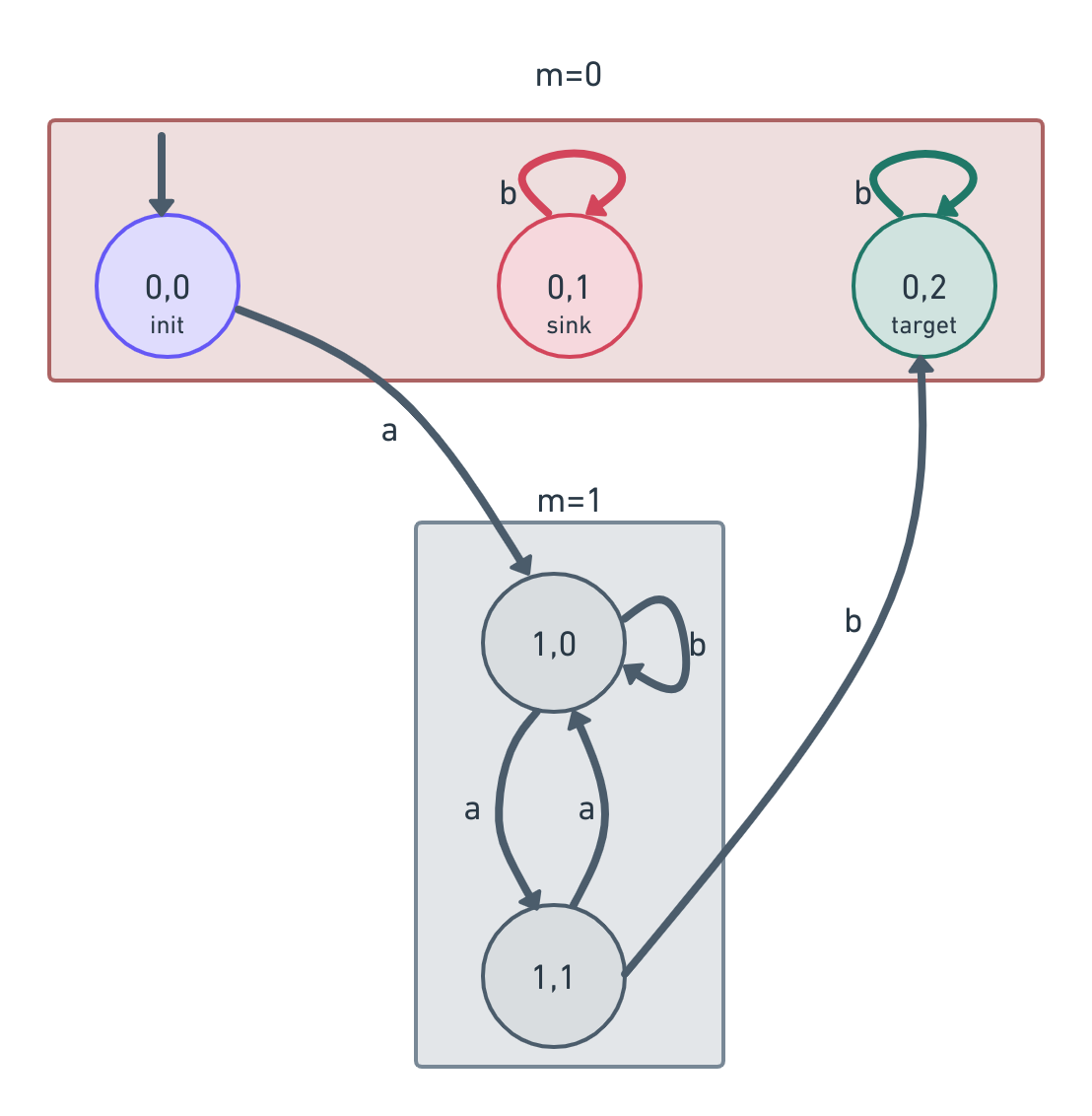}
		
		\caption{}\label{fig:modules-mdp-inst1}
	\end{subfigure}
	\begin{subfigure}{0.33\textwidth}
		\centering
		\includegraphics[width=\textwidth]{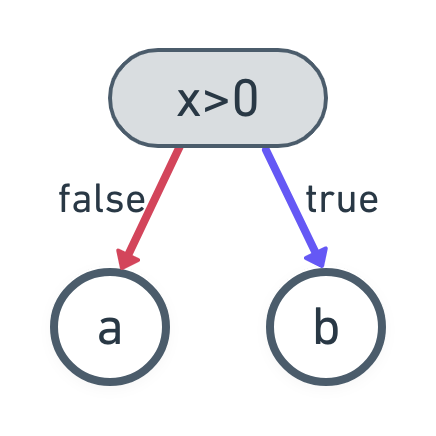}
		\caption{}
		\label{fig:dt-for-one-module-mdp}
	\end{subfigure}
	\caption{(a) MDP instance for $m=1$ from Figure~\ref{fig:modules-mdp}. (b) The optimal policy represented as a DT as learned by our approach.}
	\label{fig:modules-mdp-with-dt}
\end{figure}

In principle, one can use any of the solvable instances as the set of base instances.
However, too small instances may not contain enough information to learn a good generalization.
Therefore, we select a small set of instances $\baseinstances=\{b_1,...,b_n\}$,
such that the computed policies of these instances are rich enough to learn a generalized DT (see \cref{sec:dt-learning-details} for the details how we practically choose this set). 
This process can be seen as hyper-parameter search. 
Domain knowledge can be very useful to obtain this small set of instances.
If not available, we can choose a time budget and solve as many instances as possible in that time.

For the remainder of this section, we assume that we are given a set $\baseinstances=\{b_1,...,b_n\}$ of base instances to learn on. 
We shall interchangeably use $b_i$ for $i$th instance of the MDP, as well as for the parameters of the $i$th instance.
Hence each $b_i$ can be seen as a vector of parameters with its corresponding values. 
Formally, $b_i:=\langle p_1$=$v_1,...,$ $p_m$=$v_m\rangle$, where $p_1,...,p_m$ are the parameters and $v_1,...v_m$ are the values that are assigned to each of the respective parameters to obtain the instance.
\begin{example}
	Our running example has one parameter $k$, the number of modules. 
	In fact, we will see that it suffices to consider $\baseinstances=\{b_1\}$ where $b_1:=\langle k$=1$\rangle$, i.e.\ only learn on the simplest instance of the MDP as depicted in Figure~\ref{fig:modules-mdp-inst1}.
\end{example}

\subsection{Collecting policies}\label{sec:policies}
\begin{algorithm}[t]
	\begin{algorithmic}
		\Require A parameterized MDP $\MDP$ and base instances $\mathcal{B}=\{b_1,...,b_n\}$
		\Ensure A data set
		$D\subseteq\states_{\baseinstances} \times 2^{\actions_\baseinstances}$
		\State $D\gets \emptyset$
		\ForAll {$b_i\in \baseinstances$}
		\State Solve $\MDP_{b_i}$ and get optimal policy $\policy_{b_i}$ 
		\ForAll{$s\in \states_{b_i}$}
		\If{$s$ is reachable in $\MDP_{b_i}^{\policy_{b_i}}$ and $s\notin\gStates_{b_i}$}
		\State add $(s, \sigma_{b_i}(s))$ to $D$
		\Comment{Add optimal actions}
		\EndIf
		\EndFor
		\EndFor
		\Return {$D$};
	\end{algorithmic}
	\caption{Computing input for the DT learning from small problem instances.}\label{alg:get-table}
\end{algorithm}
We collect the optimal decisions from the optimal policies of each of the base instances into a single dataset, later to be used for learning their generalization.
The input of the learning algorithms
is a data set (possibly a multiset) of samples of (input, output)-pairs.
In our case, it is a set of pairs of the structured state and the chosen action.

Algorithm~\ref{alg:get-table} describes how to obtain the input function that can be used for the DT learning from a parameterized MDP and a set of parameterizations.
Let $\MDP_b = (\states_b,\actions_b,\trans_b,\initstate_b,\gStates_b)$ be a concrete instance of a parameterized MDP with $b:=\langle p_1$=$v_1,...,$ $p_m$=$v_m\rangle$. 
For a set of base instances $\baseinstances=\{b_1,...,b_n\}$, we denote the union of all state spaces as $\states_\baseinstances \eqdef \bigcup_{i=1}^n \states_{b_i}$, and similarly define $\actions_\baseinstances$ as the union of all action spaces.
Note that when aggregating the information, we exclude states that are not reachable in the Markov Chain (MC) induced by the computed optimal policy, as well as goal states. 
This reduces the input size for the DT learning which has two advantages: firstly, it speeds up the computation and secondly, it allows the DT to focus on the relevant states.

\begin{example}\label{ex:dt-input}
	For our running example in \cref{ex:modules-mdp}, we only consider one base instance.
	Thus, our data is given by the function $\policy_1$ for all reachable non-goal states in the MDP. 
	Concretely, we have pairs of state $(0,0)$ with action $a$, then $(1,0)$ also with $a$, and $(1,1)$ with $b$.
\end{example}

Note that the algorithms are even able to deal with data where the output is non-deterministic (different outputs for the same input within the data set).
This corresponds to accommodating \emph{permissive} policies, i.e.\ ones that allow multiple actions per state.
Thus, we can include all optimal actions from a base instance in the data set.
Similarly, we can aggregate the information over multiple base instances by a simple union of the recommended actions over all policies.
In such cases, the derived DT would also predict multiple actions.
Various determinization approaches (see \cite{AshokJJKWZ20a}) exist to get one action in these cases, e.g.,
determinization by voting picks the action that appears most often in the base instances.

\paragraph{Action labels.} 
Usually, actions are internally represented as integer values in the model-checkers, e.g. in \textsf{STORM}~\cite{storm} and \texttt{PRISM}~\cite{Prism1}. 
A natural choice would be to use these integers as labels for the actions during decision tree learning.
However, this creates a problem: When the set of actions varies among different instances, ``identical'' action choices are denoted by different integers in each instance.
\begin{example}\label{ex:dining-philosophers-label}
	Consider the dining philosophers problem in \cref{ex:dining-philosophers}. For three philosophers,
	the initial state would have $6$ actions. 
	The first $3$ are labeled with $0,1,2$, where an action $i$ represents that the $(i+1)^{th}$ philosopher thinks. 
	The second $3$ actions are labeled with $3,4,5$, where an action $i$ represents that the $(i-2)^{th}$ philosopher eats. 
	If we try to generalize this to an MDP with $n>3$ philosophers, the label $3$ is now interpreted as the action that the fourth philosopher thinks, 
	not the first philosopher eats as it was in the case of three philosophers.
	Thus, representing actions only by the index in which they appear can be sufficient if the number of modules does not change, 
	but is problematic in the opposite case.
\end{example}
To overcome this issue, we take advantage of the action-labeling feature in the \texttt{PRISM} language. An action in \texttt{PRISM} is described by a command of the following form:
\texttt{[label] guard -> prob\_1 : update\_1 +\ldots + prob\_n : update\_n}. 
This means that when the condition in the \texttt{guard} is true, \texttt{update\_i} happens with probability \texttt{prob\_i}. 
The label of the action is optional (except when the action is a synchronizing action). 
But, as a simple preprocessing step, we always define the label in each command in the \texttt{PRISM} file. 
The DT learning then can use these labels instead of the action indices. 
These labels need to be unique: assigning same label to two actions in two different modules would force the modules to take these actions simultaneously (i.e. to synchronize) changing the structure of the MDP.
Also, 
we only need to define labels for non-synchronizing actions as synchronizing actions already have labels defined that we can use.
 
For example, the problem in \cref{ex:dining-philosophers-label} can be avoided by giving the unique label \texttt{phil\_i\_line\_j} to the action for $(i+1)^{th}$ philosopher defined by the command at line $j$ in the \texttt{PRISM} file.
\subsection{Decision tree learning}

We use standard DT learning algorithms to learn a DT from the dataset constructed in the~\cref{alg:get-table}.
For predicates to be used, we consider axis-aligned predicates 
(i.e. predicates of the form $x > c$ where $x$ is a state variable
and $c\in \mathbb{R}$).
The best predicate is selected by calculating the Gini index.
Instead of having a stopping criterion,
 we let the recursive splitting of the dataset 
happen until no further splitting is possible.
The resulting DT generalizes the policy in two ways:
\begin{enumerate}
	\item 
	The DT is trained using smaller base instances. 
	The same state variables in the DT's inner node predicates are present in the larger MDP instances, but they can have a larger domain. 
	Despite this difference, the DT would still partition the state space of the larger MDP instances and still recommend actions corresponding to each state.
	\item As we are aggregating multiple policies, in our dataset, unlike the learning algorithm described in \cref{sec:prelim-dt}, we can have a state with more than one suggested actions.
	The learning algorithm considers them 
	as distinct data-points sharing the same value but different labels. 
	Since the values are the same, there are no predicates that can distinguish them. So
	these data-points traverse
	the same path in the tree until they reach a leaf node.
	The classification at the leaf node is determined by `majority voting', the label that appears most often is assigned to the leaf node. 
	This approach helps filter out actions suggested by only a few less generalizing base instances.
\end{enumerate}

\begin{example}
	For our example data set $D$ constructed in Example~\ref{ex:dt-input}, the result of the DT learning is the DT depicted in Figure~\ref{fig:dt-for-one-module-mdp}.
	This policy is in fact optimal for all $k$; see Example~\ref{ex:strat-explanation} for an explanation of this.
	In addition to being optimal, it is also small and perfectly explainable.
	
	In contrast, if we are interested in a huge instance of this model, e.g.,\ setting $k=10^{15}$, already storing the resulting MDP in the memory in order to compute an optimal policy is challenging or even infeasible for a large enough $k$. 
	Additionally, the policy produced by state-of-the-art model checkers is represented as a lookup table with as many rows as there are states.
\end{example}

\subsection{Applying and evaluating the resulting policy}\label{sec:evaluating-policies}

Once we have a decision-tree representation of a policy,
we can apply it to MDP instances of arbitrary size.
To evaluate a policy, we simply need to compute the value of the MC induced by applying the DT. 
Since solving MCs is computationally easier than solving MDPs,
we can explicitly compute values for larger MDPs (which we could not do otherwise).
Nonetheless, one can still scale the parameter to such an extent that
the construction of the corresponding MC requires too much time or memory.
In such cases, we can use statistical model checking methods~\cite{Younes02}. 

The resulting value is not only a measure for the performance of the DT policy,
but also a guaranteed lower bound on the value of the MDP (or an upper bound in the case of minimization).

Since our approach is based on generalization,
the learned DT may, in principle, recommend an action that is not available in that state.
In such cases our implementation would choose an action uniformly from the available actions.
While this may occur in principle,
we have not encountered this situation in any of the experiments we conducted for evaluation.

\section{Evaluation}
\label{sec:eval}

\subsection{Experiment Setup}
\label{sec:experiments}
\paragraph{Benchmark Selection}
We selected parameterized MDPs with reachability objective from the quantitative verification benchmark set (QVBS)~\cite{QVBS}. 
Models with reward-bounded reachability (\textit{e.g.}, eajs and resource-gathering) were excluded. 
We also identified \emph{trivial} model and property combinations
where the equation
$\min_{\policy}\prob^\policy_{\initstate} [\lozenge \gStates] = \max_{\policy}\prob^\policy_{\initstate} [\lozenge \gStates]$
holds for the set of goal states $\gStates$ and the initial state $\initstate$.
In such cases, any valid policy would act as an optimal policy.
We have excluded these from our benchmark set.
We extended the benchmark set with the Mars Exploration Rovers (mer) case study,
which was introduced in~\cite{FKP11} and appears frequently in recent literature.
This model is interesting because the probability of its property is non-trivial and
it is scalable to large parameter values without degenerating into a trivial model.

\paragraph{Choice of Base Instances}
\label{sec:dt-learning-details}
We conducted experiments to observe the effect of the set of base instances
on the value produced by the learned policy.
We synthesized decision trees from different sets of base instances, increasing the parameter(s) linearly as well as exponentially,
and evaluated them on models larger than the base instances.
We observed that one or two instances are often already enough to generalize the policy in the considered benchmark set (See \cref{sec:app-base-instances} for the chosen set of base instances used in our experiments).

\paragraph{System Configuration}
The experiments were executed on an AMD EPYC\textsuperscript{\texttrademark} 7443 server with 48 physical cores,
192 GB RAM,
running Ubuntu 22.04.2 LTS operating system with Linux kernel version 5.15.0-83-generic.
This powerful server was used to execute many runs in parallel.
We assigned 2 cores and 8 GB RAM to each run.
For all experiments, we used \benchexec~\cite{DBLP:journals/sttt/BeyerLW19}, a state of the art benchmarking tool, to isolate the executions and enforce the resource limitations.

\paragraph{Implementation Details}
We implemented our approach as an extension of the probabilistic model checker \storm~\cite{storm}.
Our code is publicly available at: \url{https://github.com/muqsit-azeem/dtstrat-123go-artifact/}.

\paragraph{Method of Comparison}
Our aim is to provide a method for policy synthesis for arbitrarily large instances of parameterized MDPs,
in particular for MDPs beyond the reach of any available rigorous analysis.
Consequently, the optimal value for such an MDP is by definition \emph{unknown} and \textbf{optimality becomes not only uncheckable,
but also unexpectable}---rather, one can hope for values close to the range where the unknown optimum is expected to lie.

Hence a straightforward evaluation is thus beyond reach, and we devise the following ancillary evaluation process.
First, we also compare on \emph{small} benchmarks,
although our approach is by no means meant as a competitor of \storm on them.
Nonetheless, it gives us the following two details:
(i) optimal values for various parameter instantiations, often allowing for a simple \emph{extrapolation},
and (ii) our relative error for these various parameter instantiations, allowing for another extrapolation.
\emph{While the performance on small models is irrelevant (of course, exact methods are to be used when feasible),
the resulting extrapolations give us some idea how our approach performs in the area of interest.}
In addition to comparing to the theoretical optimum (obtained by extrapolation),
we compare to \emph{SMC}, which is the key state-of-the-art technique for too large systems,
and to randomly chosen policies as a baseline.

\paragraph{Technical Description}
First, to obtain \emph{optimal} values for each of the MDP instances,
we executed all the engines of \storm (sparse, dd, hybrid, dd-to-sparse).
We executed each run with a CPU time limit of 1 hour and memory limit of \SI{8}{GB}.
We considered the CPU time taken by the fastest engine for each instance.

Second, we obtain values by using the state-of-the-art statistical model checker \modes~\cite{DBLP:conf/tacas/BuddeDHS18}, part of the \modest toolset~\cite{DBLP:conf/fdl/Hartmanns12}.
The approaches for picking the policies are  
(i) smart lightweight scheduler sampling (Smart LSS)~\cite{DArgenioLST15}, executed in the default configuration, 
 producing the policy value with confidence bound $0.99$ and error bound $0.01$;
(ii) the uniform policy, which resolves each non-deterministic choice by picking an action uniformly at random,
again with confidence bound $0.99$ and error bound $0.01$;
and (iii) an aggregate of 1000 randomly generated deterministic policies, i.e., non-randomizing policies where each non-deterministic choice is resolved by a single action, sampled independently according to the uniform distribution,
each evaluated with 1000 simulation runs.

Finally, we evaluate our approach by computing the value of the MCs resulting from applying our generalizing DTs.
In most of the cases (except 4), we were able to evaluate our policy precisely.
In the 4 remaining cases, we used our own implementation of SMC to evaluate the learned DT.
For 3 out of these 4, we were able to produce a value with 
with confidence bound $0.99$ and error bound $0.01$,
and in the remaining one (csma+some\_before, $N=8$) we had to use the confidence bound $0.95$ and error bound $0.05$.
The key idea of the evaluation is to show how the values (optimal / for our approach / for different random schedulers) evolve with the parameter.

We executed all the tools on the MDPs obtained by scaling the value of the parameter.
In case of MDPs with a single parameter,
we start with the smallest parameter values suggested in the QVBS and then increase it.
In cases where there was more than one parameter,
we scaled each parameter while fixing the values of the other ones.
The values chosen to be fixed were the smallest values for these parameters taken from 
the QVBS website.
Since we could not run experiments for all the parameter values due to resource constraints,
we sampled the parameters.
(See \Cref{table:param:values} in \cref{app:param-values} for the concrete parameter values used in our experiments.)

We present the results for the parameters that \storm was able to solve within a minute of CPU time,
within one hour of CPU time,
and an instance that even \storm was not able to solve within an hour of CPU time.

Sometimes, parameter scaling does not increase the time required to solve the given MDP.
In such cases, we still present several parameter valuations and the corresponding values to assess how the values evolve.

\begin{table}[t!]
	\caption{The results table for \emph{minimizing} properties. 
		\modes sometimes gives smaller value than optimal value (marked by $\sharp$)
		as it uses SMC which does not report an exact value, but reports an approximate value (with $0.01$ error bound) with high ($99\%$) confidence. For 1-2-3-Go, the values marked by $\dagger$ were approximated using SMC.
		We shorten the parameters delay to \emph{d}, deadline to \emph{dl}, and MAX\_STEPS to \emph{MS}.
		\oor\ means out-of-resources (both time and memory) and \rle\ means \emph{run length exceeded}.
		}
	\label{table:results:min}
	\begin{adjustbox}{width=1.2\textwidth,center}
			\begin{tabular}{p{0.3\textwidth}|lc|S[table-format=-1.2e-2,table-align-text-post=false] S[table-format=-1.2e-2,table-align-text-post=false] S[table-format=-1.2e-2,table-align-text-post=false] S[table-format=-1.2e-2,table-align-text-post=false] S[table-format=-1.2e-2,table-align-text-post=false] }
			\toprule
			& && \multicolumn{5}{c}{\textbf{Values}}\\
			
			Model+property & \multicolumn{2}{c|}{\textbf{Scale}} & & & \multicolumn{3}{c}{\textbf{\modes}}\\
			\cline{6-8}
			
			(values~of~parameters) 
			& Variable
			& Time
			& {\storm}
			& {1-2-3-Go}
			& {Smart LSS}
			& {Uniform}
			& {Random} \\\midrule
			
			\multirow{3}{*}{\parbox{0.3\textwidth}{zeroconf\_dl+deadline\_min\\
					(N=1000, K=1)}}
			& dl=200 &\aminute  &  \expnum{5.02e-207}  & \expnum{2.62e-43}   & 0\notemodest  & 0.0029411764705882353 & 0.0006670000000000005 \\
			& dl=1600&\anhour  &  0  &  \expnum{6.81e-86}  & 0  & 0.0012048192771084338 & 0.0006550000000000004 \\
			& dl=3200&Beyond  &  \oor  &  \expnum{8.706365528e-135}  &  0 &  0 & 0.0007180000000000005 \\
			\midrule			
			
			\multirow{3}{*}{\parbox{0.3\textwidth}{zeroconf\_dl+deadline\_min\\
					(N=1000, deadline=10)} }
			& K=2 &\aminute &  0.3388006372 & 0.3418282094 & 0.34061589679567206 & 0.3368200836820084\notemodest & 0.3422200000000006\\
			& K=8  & \aminute & 1 & 1 & 1 & 1& 1 \\[5pt]

			\midrule
			
			\multirow{3}{*}{\parbox{0.3\textwidth}{firewire+deadline\\ (deadline=200)}}
			& delay=5& \aminute & 0.5 & 0.5 & 0.5623232709209018 & 0.99453125 & 0.9909990000000001\\
			&delay=21 &\aminute  &  0.5  &  0.5  & 1  & 1 & 1 \\
			&delay=34 &\aminute  &  0  &  0  & 1  & 1 & 1 \\
			&delay=89 &\aminute  &  0  &  0  & 1  & 1 & 1 \\			
			\midrule
			
			\multirow{3}{*}{\parbox{0.3\textwidth}{firewire+deadline\\ (delay=3)}}
			& dl=200  &  \aminute & 0.5  &  0.5  & 0.4976594941487354\notemodest  & 0.9770491803278688 & 0.9709269999999999 \\
			& dl=500 & \aminute & 0.8515625 & 0.8515625 & 1 & 1 & 1\\
			& dl=1300 & \aminute& 0.9983077347 & 0.9983077347 & 1 & 1 & 1 \\
			\midrule
			
			\multirow{2}{*}{\parbox{0.3\textwidth}{csma+some\_before\\ (N=2)}}
			& K=2 & \aminute  &  0.5  & 0.5  & 0.4955849056603774\notemodest  & 0.49818799546998865\notemodest & 0.5003709999999995 \\
			& K=3 & \aminute  &  0.875  &  0.875  & 0.875  & 0.8726829268292683\notemodest & 0.8747569999999995\notemodest \\
			\midrule
			
			\multirow{3}{*}{\parbox{0.3\textwidth}{csma+some\_before\\ (fix K=2)}}
			& N=3 & \aminute  &  0.5859375  &  0.5859375  & 0.5825028968713789\notemodest &  0.8944128787878788 & 0.9025649999999997 \\
			& N=5 & \anhour  &  0.2056884766  &  0.2056884766  & 0.39026409144659047  & 0.7852764094143404 & 0.7778539999999998 \\
			& N=8 & \anhour  &  0.03542137146  &  0.03256445006 {$^{\dagger}$}
			 & 0.5096981132075472  & 0.7540229885057471 & 0.7602989999999993 \\
			& N=13 & Beyond  &  \oor  &  \oor  &  0.5638623326959847 & 0.7527390438247012 & 0.7618870000000004 \\[2mm]
			
			\hline\hline
		
			\multirow{3}{*}{\parbox{0.3\textwidth}{consensus+c2\\ (N=2)}}
			& K=2 & \aminute &  0.3828112753  &  0.46875  & 0.41573859242072697 & 0.4921509433962264 & 0.48594900000000113 \\
			& K=55  & \aminute &  0.4939189539  &  0.4954547318 & \rle  & \rle & \rle \\
			& K=144  & \aminute & 0.4879562054  &  0.4982656911  & \rle  & \rle & \rle \\
			\midrule
			
			\multirow{3}{*}{\parbox{0.3\textwidth}{consensus+c2\\ (K=2)}}
			& N=6 &\aminute  &  0.2943375674  &  0.4477979415  & 0.48901472253680633  & 0.48297470743676857 & 0.4831310000000007 \\
			& N=7 &\anhour  &  0.2877048806  &  0.4568046327 & 0.48175981873111784  & 0.4858814647036618 & 0.48219200000000145 \\
			& N=13 & Beyond  &  \oor  &  0.4647543657  &  \rle & 0.4858814647036618  &  \rle \\
			\midrule
			
			\multirow{3}{*}{\parbox{0.3\textwidth}{zeroconf\_dl+deadline\_min\\
					(deadline=10, K=1)} }
			& N=1000 &\aminute &  0.001424816451 & 0.004227527321 & 0.0029411764705882353& 0.007432432432432433\notemodest & 0.006983000000000012\\
			& N=8000  & \aminute & 0.01391008151 & 0.03848590599 & 0.03505882352941177 & 0.07304236200256739 & 0.060227000000000086\\
			& N=32000 & \aminute&0.1068034215& 0.2228617513 & 0.20887276785714284 & 0.3545380212591987 & 0.31968900000000067  \\
			\midrule

			\multirow{3}{*}{\parbox{0.3\textwidth}{pacman+crash}}
			& MS=5 &\aminute &  0.5511 & 0.5510967 & 0.5507036896158235\notemodest & 0.5494296577946768 & 0.5507580000000006\notemodest\\
			& MS=25  & \aminute & 0.5511 & 0.8693693636 & 0.7285714285714285 & 0.9271557271557271 & 0.9226340000000002 \\
			& MS=200 & \anhour & 0.5511348333 & 0.9999792265 & 1 & 1 &  1 \\
			& MS=300 & Beyond & \oor & 0.9999792265 & 1 & 1 & 1\\
			
			\bottomrule
		\end{tabular}
	\end{adjustbox}
\end{table}

\subsection{Results}
\label{sec:results}
\Cref{table:results:min} and \Cref{table:results:max} show the results of our evaluation for the minimizing and maximizing properties, respectively. 
Each \emph{instance} refers to a combination of model, property and parameter.
The tables show, for each model+property combination,
the parameter value and CPU time taken by \storm to solve it
(\aminute, \anhour, and Beyond in case \storm could not solve the instance in an hour),
the values produced by \storm, our approach and the sampling based SMC.
The tables report \oor \xspace when running out of resource (time or memory).
Also, a few \modes runs resulted in a \emph{run length exceeded} (\rle) error.

\begin{table}[t!]
	\setlength\fboxsep{1cm}
	\caption{The results table for \emph{maximizing} properties.
 The values marked by $\dagger$ were approximated using SMC.
		We shorten the parameter deadline to \emph{dl}.
		\oor \ means out-of-resources (both time and memory) and \rle \ means \emph{run length exceeded}.
		 }
	\label{table:results:max}
	\begin{adjustbox}{width=1.2\textwidth,center}	\begin{tabular}{p{0.3\textwidth}|lc|S[table-format=-1.2e-2,table-align-text-post=false] S[table-format=-1.2e-2,table-align-text-post=false] S[table-format=-1.2e-2,table-align-text-post=false] S[table-format=-1.2e-2,table-align-text-post=false] S[table-format=-1.2e-2,table-align-text-post=false] }
			\toprule
			& & & \multicolumn{5}{c}{\textbf{Values}}\\
			
			Model+property& \multicolumn{2}{c|}{\textbf{Scale}} & & & \multicolumn{3}{c}{\textbf{\modes}}\\			
			\cline{6-8}

			(values~of~parameters) 
			& Variable
			& Time
			& {\storm}
			& {1-2-3-Go}
			& {Smart LSS}
			& {Uniform}
			& {Random} \\\midrule

			\multirow{3}{*}{\parbox{0.3\textwidth}{zeroconf\_dl+deadline\_max \\	(N=1000, K=1)}}
			&	dl=200& \aminute  & 0.005397647598   &  0.005390585799  & 0.002150537634408602  & 0.0029411764705882353 & 0.0007530000000000005 \\
			& dl=1600& \anhour  &  0.005397647598  &  0.005390585799  & 0.0012048192771084338  & 0.0012048192771084338 & 0.0006550000000000004 \\
			& dl=3200& Beyond  &  \oor  &  0.005390585799  & 0.0012048192771084338  & 0 & 0.0006360000000000005 \\
			\midrule
			
			\multirow{3}{*}{\parbox{0.3\textwidth}{zeroconf\_dl+deadline\_max \\
					(N=1000, deadline=10)} }
			& K=2  & \aminute &  0.343585958 & 0.34356 & 0.34046627810158203 & 0.3429875518672199 & 0.3418300000000005\\
			& K=8  & \aminute &  1 & 1 & 1 & 1& 1 \\
			& K=32  & \aminute &  1 & 1 & 1 & 1& 1  \\
			\midrule
			
			\multirow{3}{*}{\parbox{0.3\textwidth}{zeroconf\_dl+deadline\_max  \\
					(deadline=10, K=1)} }
			& N=1000  & \aminute & 0.01537893701 & 0.01536 & 0.009826589595375723 & 0.00909090909090909 & 0.006871999999999993\\
			& N=8000   & \aminute & 0.1230314961 & 0.12303 & 0.10308880308880308 & 0.06748971193415639 & 0.06108600000000007\\
			& N=32000 & \aminute & 0.4921259843 & 0.49212 & 0.46672975018925056 & 0.3471316549731738 & 0.3197710000000013\\
			\midrule
			
			\multirow{3}{*}{\parbox{0.3\textwidth}{philosophers+eat}}
			&N=5 & \aminute  &  1  &  1  & 1  & 1 & 0.03734400000000001 \\
			&N=21& \anhour  &   1 &  1  & 0.5055094339622641  & 1 & 0.000245 \\
			&N=34& Beyond  &   \oor &   1 &  0.4948301886792453 & 1 & 0 \\
			\midrule
			
			\multirow{3}{*}{\parbox{0.3\textwidth}{pnueli-zuck+live}}
			&N=5 &\aminute  &  1  & 1   & 1  & 1 & 0.035749 \\
			&N=21 &\anhour  &  1  & 1   & 1  & 1 & 0.002 \\
			&N=34 & Beyond  &  \oor  & 1   & 1  & 1 & 0.002 \\
			\midrule			
			
			\multirow{3}{*}{\parbox{0.3\textwidth}{csma+all\_before\_max \\ (N=2)}}
			& K=8 & \aminute  &  1  & 1   & 1  & 1 & 1 \\
			& K=11 & \anhour  &  1  &  1 {$^\dagger$}  & 1  & 1 & 1 \\
			& K=13 & Beyond  &  \oor  &  1 {$^\dagger$} &  1 & 1 & \oor \\
			\midrule

			\multirow{3}{*}{\parbox{0.3\textwidth}{mer+p1 \\ (x=0.01)}}
			& n=1000& \aminute  &  0.2016129032  & 0.19999  & \rle  & \rle & 0.000184\\
			& n=21000 &\anhour  &  0.2016128  &  0.19999  & \rle  & \rle  & 0\\
			& n=55000 & Beyond  &  \oor  & 0.19999 & \rle  & \rle & 0.000399 \\[2mm]
			
			\hline\hline

			\multirow{3}{*}{\parbox{0.3\textwidth}{mer+p1 \\ (n=10)}}
			& x=0.01  &  \aminute & 0.2016128  & 0.19999  & \rle  & \rle  & 0.000402\\
			& x=0.08  & \aminute &0.213675199   &  0.19999  & \rle  & \rle & 0.000175\\
			& x=0.55  &  \aminute &0.3570457237  & 0.19999 & \rle  & \rle  & 0.000494 \\
			& x=0.89 & \aminute  &  0.6711936689  & 0.19999  & \rle  & \rle & 0.0005809999999999999\\			
			\midrule
			
			\multirow{3}{*}{\parbox{0.3\textwidth}{consensus+disagree \\ (K=2)}}
			& N=5 & \aminute  &  0.3356236942  &  0.10078469  &  0.04583333333333333 & 0.029 & 0.03502900000000011 \\
			& N=7 &\anhour  &  0.3839650052  &  0.06586586844  &  0.03308641975308642 & 0.043 & 0.03409800000000015 \\
			& N=13& Beyond  &  \oor  &  0.04586406712  &  \rle & 0.03125 & \rle \\
			\midrule			
			
			\multirow{3}{*}{\parbox{0.3\textwidth}{consensus+disagree \\ (N=2)}}
			& K=2  & \aminute& 0.1083325973  &  0.0625 & 0.075 & 0.027887323943661974 & 0.027651000000000005\\
			& K=55  &  \aminute &0.004501618143 & 0.003223781156 & \rle & \rle & \rle \\
			& K=144  &  \aminute& 0.001625395845  &  0.0004573592205  &  \rle & \rle & \rle \\
			\midrule
			
			\multirow{3}{*}{\parbox{0.3\textwidth}{csma+all\_before\_max \\ (K=2)}}
			& N=3 &\aminute  &  0.8596150365  &  0.5249656586  & 0.8506446991404012  & 0.032754342431761785 & 0.6838249999999996 \\
			& N=5 &\anhour  &  0.7006558068  &  0.04993145573  &  0.41336173508907825 & 0.0357308584686775 & 0.2363679999999993 \\
			& N=6& Beyond  &  \oor  & 0.009437167086 {$^\dagger$}  & 0.20984427141268075  & 0.03869565217391304 & 0.11185200000000006 \\
			
			\bottomrule
		\end{tabular}
	\end{adjustbox}
\end{table}

Some models converge to triviality (\textit{i.e.}, max=min) as we scale the parameter.
\emph{Zeroconf\_dl+deadline\_min} becomes trivial for higher values of the parameter $K$ than $3$,
\emph{firewire+deadline} becomes trivial for deadline $>$ 1300,
and \emph{csma+some\_before} becomes trivial when the value of $K$ is more than twice the value of $N$.
Pacman also approaches closer to triviality for higher values of the parameter MAX\_STEPS (the horizon).

The results show that our approach gives
near optimal values for 13 out of 21 cases (the upper halves of \Cref{table:results:min} and \Cref{table:results:max}),
better than Smart LSS for 2 out of remaining 8 cases,
and generally better than random and uniform in remaining cases.
There are two instances where random performs better than our approach
(pacman for the MAX\_STEPS 25, and csma+all\_before\_max for N=3),
see the discussion below.

\subsection{Discussion} 
\begin{figure}[t]
	\centering
	\ref{mylegend}\\
	\begin{subfigure}[t]{0.49\textwidth}
			\centering
			    \begin{tikzpicture}
	\begin{axis}[
		height=4cm,
		width=\textwidth,
		xlabel=deadline,
		ylabel=value,
		xmin=0, xmax=3300,
		ymin=-0.001, ymax=0.009,
		legend style={at={(1,1)}},
		ytick={0,0.004,0.008},
		scaled y ticks = false,
		y tick label style={/pgf/number format/fixed},
		yticklabels={0,0.004,0.008},
		legend to name={mylegend},
		legend style={at={(0.5,-0.2)},    
			anchor=north,legend columns=3, /tikz/every even column/.append style={column sep=0.5cm}},  
		]
\addplot+[dotted,-,mark=o,mark options={solid,scale=0.75}] plot coordinates {
	(100,0.005397647598)
	(200,0.005397647598)
	(300,0.005397647598)
	(600,0.005397647598)
	(1000,0.005397647598)
	(1600,0.005397647598)
	(2000,0.005397647598)
	(2400,0.005397647598)
	(2800,0.005397647598)
	(3200,0.005397647598)
};

\addplot+[dotted,only marks,-,mark=square,mark options={solid,scale=0.75}] plot coordinates {
	(100,0.005397647598)
	(200,0.005397647598)
	(300,0.005397647598)
	(600,0.005397647598)
	(1000,0.005397647598)
	(1600,0.005397647598)
};
		\addplot+[dotted,-,mark=star,mark options={solid,scale=0.75}] plot coordinates {
			(100,0.0012048192771084338)
			(200,0.002150537634408602)
			(300,0.0)
			(600,0.0029411764705882353)
			(1000,0.0012048192771084338)
			(1600,0.0012048192771084338)
			(2000,0.0012048192771084338)
			(2400,0.0012048192771084338)
			(2800,0.003669724770642202)
			(3200,0.0012048192771084338)
		};
\legend{1-2-3-Go!,\storm,\modes-LSS}  
	\end{axis}
\end{tikzpicture}
			\vspace{-4mm}
			\caption{zeroconf\_dl + deadline\_max}
			\label{fig:robustness:zeroconf}
		\end{subfigure}
	\hfill
	\begin{subfigure}[t]{0.49\textwidth}
			\centering
				\begin{tikzpicture}
		\begin{axis}[
			height=4cm,
			width=\textwidth,
			xlabel=$N$,
			ylabel=value,
			xmin=0, xmax=60,
			ymin=-0.2, ymax=1.2,
			legend style={at={(1,0.8)}},
			]
			\addplot+[dotted,-,mark=o,mark options={solid,scale=0.75}] plot coordinates {
				(3,1)
				(4,1)
				(5,1)
				(8,1)
				(13,1)
				(21,1)
				(24,1)
				(27,1)
				(30,1)
				(34,1)
				(55,1)
			};
			
			\addplot+[dotted,only marks,-,mark=square,mark options={solid,scale=0.75}] plot coordinates {
				(3,1)
				(4,1)
				(5,1)
				(8,1)
			};
			
			\addplot+[dotted,-,mark=star,mark options={solid,scale=0.75}] plot coordinates {
				(3,1)
				(4,1)
				(5,1)
				(8,1)
				(13,0.5012080030200076)
				(21,0.5055094339622641)
				(24,0.4960754716981132)
				(27,0.06943699731903485)
				(30,0.500641751604379)
				(34,0.4948301886792453)
				(55,0)
			};
			
		\end{axis}
	\end{tikzpicture}
			\caption{philosophers + eat}
			\label{fig:robustness:phil}
		\end{subfigure}
	\caption{Robustness of the policy given by 1-2-3-Go!, \storm and
			\modes for different instances of zeroconf\_dl + deadline\_max and philosophers + eat. For larger models, while \storm times out and \modes gives sub-optimal policies, 1-2-3-Go! consistently gives better results irrespective of $N$.
		}
	\label{fig:value-evolution}
\end{figure}
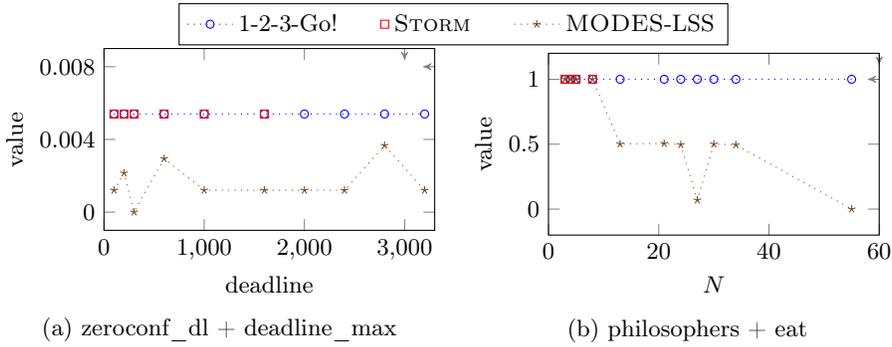

Although our approach is simple,
it performs well in a number of cases.
We often generalize from a single instance or two, yielding satisfactory solutions for arbitrarily large instantiations.
In a number of cases, we can justifiably extrapolate that the policies are (nearly) optimal for all instances.
For instance, consider the two benchmarks of Figure~\ref{fig:value-evolution}.
No matter how much the model is scaled up, the value of our policy seems to remain stable.
While its (near-)optimality can be proven only up to a certain point (beyond which no ground truth can be known), the apparent stability suggests it is true onwards, too.
Note that \modes returns low values as the optimal policies are rather rare.

In the sequel, we discuss the scope and the limitations of our approach in details. As discussed earlier, we can divide the parameters in two types.

\paragraph{Type 1:} Parameters that dictate the number of \prism-modules. 
This type of parameter not only changes the structure of the MDP, but also increases the number of state variables. 
Note that when we train a DT from the policies from smaller base instances, 
the predicates in decision tree would not use the state variables present only in the bigger instances. 

Even then, as the system is a product of these isomorphic modules, we can think the smaller base instances as \emph{projections} of the larger instance to the first few state variables.
Then, for some interesting properties, our method still gives good policies: cases where there is a generalizing optimal policy that does not depend on the additional modules.

For example, consider the \emph{csma} model describing the CSMA/CD consensus protocol when $N$ stations use a network with a single channel. 
Each station is represented by a module in the \prism file. 
Now consider two different properties ``all\_before\_max'' and ``some\_before''. the first one checks the maximum probability that all stations send the message successfully avoiding a data collision.
A policy maximizing successful transmission for all stations, needs to take account of the state variables corresponding to all modules in the \prism file. 
Our approach fails with this kind of property.
But on the other hand, the second property checks the minimum probability that some station eventually sends the message successfully.
Thus, a decision tree generated from the base instances gives a policy that minimizes the collision probability for some station among the first three stations. 
This would act as an optimal policy (as it optimizes the collision probability for some station) 
in the larger instances even though the predicates in the DT does not contain state variables related to stations with larger index.
Thus, for this property, our approach succeeds in generalizing the optimal policy.

\paragraph{Type 2 :} Other parameters which can be changed by setting the value externally. 
Often, this would mean expanding the domain of the state variables (as in the case of \Cref{ex:modules-mdp}), which increases the size of the state space linearly.
Then our approach can still work as we can have an optimal policy that is independent of that specific state variable, or if the newly added states are not relevant. 

However, this is not always the case: 
The parameter MAX\_STEPS in \emph{pacman + crash} denotes the number of steps Pac-Man needs to stay safe from the ghosts. 
Our algorithm fails for this model as the policies across instances cannot be generalized. 
Indeed, a policy that minimized the probability of crash for $K$ steps does not provide any information about how to stay safe for $K'>K$ steps. 

In the case of \emph{mer + p1}, if we fix $n$ and vary the parameter $x$, our approach fails to generalize a policy. 
Changing the value of $x$ does not change the state-space or the structure of the model, 
but changes the probability values of the transitions, 
which in turn would change the optimal policy across different instances. 
For that reason, we cannot construct a generalizing DT as the decision predicates are defined on the state variables 
and not on the probability value of the transition.

\paragraph{Time}
The approach typically needs less than \SI{5}{s} to generate a policy, except for zeroconf\_dl (see \cref{table:results:min:time} and \cref{table:results:max:time} in \cref{app:time}).
In the case of zeroconf\_dl, the time taken is a couple of minutes because we learned from a larger instance, with non-trivial values for all three involved parameters.
However, this instance is then so informative that we could later use it for all instances derived by varying different parameters
(i.e., varying N and fixing others, varying delay and fixing other, and varying deadline and fixing others, not having to take care of these three families separately).

\section{Conclusion}
We have seen that \emph{practically good} policies (with values close to the unknown optimum in the sense of the ``Method of Comparison'' above) can be generated in a lightweight way even for very large parameterized models, beyond reach of any other methods.
In order to synthesize policies for arbitrarily large models,
we generalize the policies computed for the smaller instances using (more explainable and thus more generalizable) decision trees, coining the ``generalizability by explainability''. 

The generalization is an example of \emph{unreliable} reasoning, which can contribute to better scalability.
On the one hand,
the unreliability results in \emph{no guarantees} that the produced policies are anywhere close to optimum,
which, however, often cannot be computed anyway.
On the other hand, the values of the policies can be \emph{reliably} approximated:
either numerically with absolute guarantees if the resulting Markov chain is still analyzable (e.g.,\ with partial-exploration methods \cite{DBLP:journals/lmcs/KretinskyM20}) or with statistical guarantees by SMC on the Markov chain. 
Consequently, although optimal control policies might be out of reach,
we can still produce what we thus coin here as \emph{provably good enough} policies.
Moreover, the consistency of the values over the different instantiations often suggests practical proximity to optimum.

A possibly surprising point is the conclusion of our experiments that very few base instances need to be analyzed.
Such \emph{robustness} (together with the robustness across the target instances as seen in  \cref{fig:value-evolution}) suggests that this generalizability is a deeply inherent property of many models, and thus deserves further investigation and exploitation.
In particular, our approach is only the first, generic try to exploit this property, opening the new paradigm.
As suggested by our experimental results,
more \emph{specific heuristics} for certain types of systems where parameters play different roles,
such as number of modules, number of repetitions, time-outs, etc.,
offer a desirable direction of future work.

\bibliographystyle{splncs04}
\bibliography{ref}

\begin{thebibliography}{10}
\providecommand{\url}[1]{\texttt{#1}}
\providecommand{\urlprefix}{URL }
\providecommand{\doi}[1]{https://doi.org/#1}

\bibitem{AshokJJKWZ20a}
Ashok, P., Jackermeier, M., Jagtap, P., Kret{\'{\i}}nsk{\'{y}}, J., Weininger,
  M., Zamani, M.: dtcontrol: decision tree learning algorithms for controller
  representation. In: Ames, A.D., Seshia, S.A., Deshmukh, J. (eds.) {HSCC} '20:
  23rd {ACM} International Conference on Hybrid Systems: Computation and
  Control, Sydney, New South Wales, Australia, April 21-24, 2020. pp.
  30:1--30:2. {ACM} (2020). \doi{10.1145/3365365.3383468},
  \url{https://doi.org/10.1145/3365365.3383468}

\bibitem{dtcontrol2}
Ashok, P., Jackermeier, M., Kret{\'{\i}}nsk{\'{y}}, J., Weinhuber, C.,
  Weininger, M., Yadav, M.: dtcontrol 2.0: Explainable strategy representation
  via decision tree learning steered by experts. In: {TACAS} {(2)}. Lecture
  Notes in Computer Science, vol. 12652, pp. 326--345. Springer (2021)

\bibitem{BK08}
Baier, C., Katoen, J.: Principles of model checking. {MIT} Press (2008)

\bibitem{DBLP:conf/cav/Baier0L0W17}
Baier, C., Klein, J., Leuschner, L., Parker, D., Wunderlich, S.: Ensuring the
  reliability of your model checker: Interval iteration for markov decision
  processes. In: {CAV} {(1)}. Lecture Notes in Computer Science, vol. 10426,
  pp. 160--180. Springer (2017)

\bibitem{DBLP:journals/sttt/BeyerLW19}
Beyer, D., L{\"{o}}we, S., Wendler, P.: Reliable benchmarking: requirements and
  solutions. Int. J. Softw. Tools Technol. Transf.  \textbf{21}(1),  1--29
  (2019). \doi{10.1007/s10009-017-0469-y},
  \url{https://doi.org/10.1007/s10009-017-0469-y}

\bibitem{BrazdilCCFK15}
Br{\'{a}}zdil, T., Chatterjee, K., Chmelik, M., Fellner, A.,
  Kret{\'{\i}}nsk{\'{y}}, J.: Counterexample explanation by learning small
  strategies in markov decision processes. In: Kroening, D., Pasareanu, C.S.
  (eds.) Computer Aided Verification - 27th International Conference, {CAV}
  2015, San Francisco, CA, USA, July 18-24, 2015, Proceedings, Part {I}.
  Lecture Notes in Computer Science, vol.~9206, pp. 158--177. Springer (2015).
  \doi{10.1007/978-3-319-21690-4\_10},
  \url{https://doi.org/10.1007/978-3-319-21690-4\_10}

\bibitem{atva14}
Br{\'{a}}zdil, T., Chatterjee, K., Chmelik, M., Forejt, V.,
  Kret{\'{\i}}nsk{\'{y}}, J., Kwiatkowska, M.Z., Parker, D., Ujma, M.:
  Verification of markov decision processes using learning algorithms. In:
  {ATVA}. Lecture Notes in Computer Science, vol.~8837, pp. 98--114. Springer
  (2014)

\bibitem{breiman1984classification}
Breiman, L.: Classification and Regression Trees. (The Wadsworth statistics /
  probability series), Wadsworth International Group (1984)

\bibitem{DBLP:conf/tacas/BuddeDHS18}
Budde, C.E., D'Argenio, P.R., Hartmanns, A., Sedwards, S.: A statistical model
  checker for nondeterminism and rare events. In: {TACAS} {(2)}. Lecture Notes
  in Computer Science, vol. 10806, pp. 340--358. Springer (2018).
  \doi{10.1007/978-3-319-89963-3\_20},
  \url{https://doi.org/10.1007/978-3-319-89963-3\_20}

\bibitem{QComp}
Budde, C.E., Hartmanns, A., Klauck, M., Kret{\'{\i}}nsk{\'{y}}, J., Parker, D.,
  Quatmann, T., Turrini, A., Zhang, Z.: On correctness, precision, and
  performance in quantitative verification - {QComp} 2020 competition report.
  In: ISoLA {(4)}. Lecture Notes in Computer Science, vol. 12479, pp. 216--241.
  Springer (2020)

\bibitem{CiesinskiBGK08}
Ciesinski, F., Baier, C., Gr{\"{o}}{\ss}er, M., Klein, J.: Reduction techniques
  for model checking markov decision processes. In: {QEST}. pp. 45--54. {IEEE}
  Computer Society (2008)

\bibitem{DArgenioLST15}
D'Argenio, P.R., Legay, A., Sedwards, S., Traonouez, L.: Smart sampling for
  lightweight verification of markov decision processes. Int. J. Softw. Tools
  Technol. Transf.  \textbf{17}(4),  469--484 (2015)

\bibitem{Feng14}
Feng, L.: On learning assumptions for compositional verification of
  probabilistic systems. Ph.D. thesis, University of Oxford, {UK} (2014)

\bibitem{FKP11}
Feng, L., Kwiatkowska, M., Parker, D.: Automated learning of probabilistic
  assumptions for compositional reasoning. In: Giannakopoulou, D., Orejas, F.
  (eds.) Fundamental Approaches to Software Engineering. pp. 2--17. Springer
  Berlin Heidelberg, Berlin, Heidelberg (2011)

\bibitem{GrooteVV18}
Groote, J.F., Verduzco, J.R., de~Vink, E.P.: An efficient algorithm to
  determine probabilistic bisimulation. Algorithms  \textbf{11}(9), ~131 (2018)

\bibitem{GrosH0KS20}
Gros, T.P., Hermanns, H., Hoffmann, J., Klauck, M., Steinmetz, M.: Deep
  statistical model checking. In: {FORTE}. Lecture Notes in Computer Science,
  vol. 12136, pp. 96--114. Springer (2020)

\bibitem{DBLP:conf/fmco/GroesserB05}
Gr{\"{o}}{\ss}er, M., Baier, C.: Partial order reduction for markov decision
  processes: {A} survey. In: {FMCO}. Lecture Notes in Computer Science,
  vol.~4111, pp. 408--427. Springer (2005)

\bibitem{DBLP:conf/rp/HaddadM14}
Haddad, S., Monmege, B.: Reachability in mdps: Refining convergence of value
  iteration. In: {RP}. Lecture Notes in Computer Science, vol.~8762, pp.
  125--137. Springer (2014)

\bibitem{DBLP:conf/tacas/HahnPSSTW19}
Hahn, E.M., Perez, M., Schewe, S., Somenzi, F., Trivedi, A., Wojtczak, D.:
  Omega-regular objectives in model-free reinforcement learning. In: {TACAS}
  {(1)}. Lecture Notes in Computer Science, vol. 11427, pp. 395--412. Springer
  (2019)

\bibitem{DBLP:conf/fdl/Hartmanns12}
Hartmanns, A.: {MODEST} - {A} unified language for quantitative models. In:
  {FDL}. pp. 44--51. {IEEE} (2012),
  \url{https://ieeexplore.ieee.org/document/6336982/}

\bibitem{HH15}
Hartmanns, A., Hermanns, H.: Explicit model checking of very large {MDP} using
  partitioning and secondary storage. In: Finkbeiner, B., Pu, G., Zhang, L.
  (eds.) Automated Technology for Verification and Analysis - 13th
  International Symposium, {ATVA} 2015, Shanghai, China, October 12-15, 2015,
  Proceedings. Lecture Notes in Computer Science, vol.~9364, pp. 131--147.
  Springer (2015). \doi{10.1007/978-3-319-24953-7\_10},
  \url{https://doi.org/10.1007/978-3-319-24953-7\_10}

\bibitem{QVBS}
Hartmanns, A., Klauck, M., Parker, D., Quatmann, T., Ruijters, E.: The
  quantitative verification benchmark set. In: {TACAS} {(1)}. Lecture Notes in
  Computer Science, vol. 11427, pp. 344--350. Springer (2019).
  \doi{10.1007/978-3-030-17462-0\_20},
  \url{https://doi.org/10.1007/978-3-030-17462-0\_20}

\bibitem{HartmannsT15}
Hartmanns, A., Timmer, M.: Sound statistical model checking for {MDP} using
  partial order and confluence reduction. Int. J. Softw. Tools Technol. Transf.
   \textbf{17}(4),  429--456 (2015)

\bibitem{DBLP:conf/qest/HenriquesMZPC12}
Henriques, D., Martins, J.G., Zuliani, P., Platzer, A., Clarke, E.M.:
  Statistical model checking for markov decision processes. In: {QEST}. pp.
  84--93. {IEEE} Computer Society (2012)

\bibitem{storm}
Hensel, C., Junges, S., Katoen, J., Quatmann, T., Volk, M.: The probabilistic
  model checker storm. Int. J. Softw. Tools Technol. Transf.  \textbf{24}(4),
  589--610 (2022). \doi{10.1007/s10009-021-00633-z},
  \url{https://doi.org/10.1007/s10009-021-00633-z}

\bibitem{HYAFIL197615}
Hyafil, L., Rivest, R.L.: Constructing optimal binary decision trees is
  np-complete. Information Processing Letters  \textbf{5}(1),  15--17 (1976).
  \doi{https://doi.org/10.1016/0020-0190(76)90095-8},
  \url{https://www.sciencedirect.com/science/article/pii/0020019076900958}

\bibitem{Kamaleson18}
Kamaleson, N.: Model reduction techniques for probabilistic verification of
  Markov chains. Ph.D. thesis, University of Birmingham, {UK} (2018)

\bibitem{KleinBCDDKMM18}
Klein, J., Baier, C., Chrszon, P., Daum, M., Dubslaff, C., Kl{\"{u}}ppelholz,
  S., M{\"{a}}rcker, S., M{\"{u}}ller, D.: Advances in probabilistic model
  checking with {PRISM:} variable reordering, quantiles and weak deterministic
  b{\"{u}}chi automata. Int. J. Softw. Tools Technol. Transf.  \textbf{20}(2),
  179--194 (2018)

\bibitem{DBLP:journals/lmcs/KretinskyM20}
Kret{\'{\i}}nsk{\'{y}}, J., Meggendorfer, T.: Of cores: {A} partial-exploration
  framework for markov decision processes. Log. Methods Comput. Sci.
  \textbf{16}(4) (2020)

\bibitem{Prism1}
Kwiatkowska, M.Z., Norman, G., Parker, D.: {PRISM:} probabilistic symbolic
  model checker. In: Computer Performance Evaluation / {TOOLS}. Lecture Notes
  in Computer Science, vol.~2324, pp. 200--204. Springer (2002)

\bibitem{DBLP:conf/qest/KwiatkowskaNP06}
Kwiatkowska, M.Z., Norman, G., Parker, D.: Game-based abstraction for markov
  decision processes. In: {QEST}. pp. 157--166. {IEEE} Computer Society (2006)

\bibitem{KwiatkowskaNP06}
Kwiatkowska, M.Z., Norman, G., Parker, D.: Symmetry reduction for probabilistic
  model checking. In: {CAV}. Lecture Notes in Computer Science, vol.~4144, pp.
  234--248. Springer (2006)

\bibitem{PRISMben}
Kwiatkowska, M.Z., Norman, G., Parker, D.: The {PRISM} benchmark suite. In:
  {QEST}. pp. 203--204. {IEEE} Computer Society (2012).
  \doi{10.1109/QEST.2012.14}, \url{https://doi.org/10.1109/QEST.2012.14}

\bibitem{KwiatkowskaPQ11}
Kwiatkowska, M.Z., Parker, D., Qu, H.: Incremental quantitative verification
  for markov decision processes. In: {DSN}. pp. 359--370. {IEEE} Compute
  Society (2011)

\bibitem{li2019compositional}
Li, R., Liu, Y.: Compositional stochastic model checking probabilistic automata
  via symmetric assume-guarantee rule. In: 2019 IEEE 17th International
  Conference on Software Engineering Research, Management and Applications
  (SERA). pp. 110--115. IEEE (2019)

\bibitem{LomuscioP19}
Lomuscio, A., Pirovano, E.: A counter abstraction technique for the
  verification of probabilistic swarm systems. In: {AAMAS}. pp. 161--169.
  International Foundation for Autonomous Agents and Multiagent Systems (2019)

\bibitem{maisonneuve2009automatic}
Maisonneuve, V.: Automatic heuristic-based generation of mtbdd variable
  orderings for prism models. internship report (2009)

\bibitem{mitchell1997machine}
Mitchell, T.: Machine learning, vol.~1. McGraw-hill New York (1997)

\bibitem{MohagheghiS20}
Mohagheghi, M., Salehi, K.: Machine learning and disk-based methods for
  qualitative verification of markov decision processes. In: {ICTERI}
  Workshops. {CEUR} Workshop Proceedings, vol.~2732, pp. 74--88. CEUR-WS.org
  (2020)

\bibitem{Parker03}
Parker, D.A.: Implementation of symbolic model checking for probabilistic
  systems. Ph.D. thesis, University of Birmingham, {UK} (2003)

\bibitem{DBLP:books/wi/Puterman94}
Puterman, M.L.: Markov Decision Processes: Discrete Stochastic Dynamic
  Programming. Wiley Series in Probability and Statistics, Wiley (1994).
  \doi{10.1002/9780470316887}, \url{https://doi.org/10.1002/9780470316887}

\bibitem{PH99}
Pyeatt, L.D., Howe, A.E.: Decision tree function approximation in reinforcement
  learning (1999)

\bibitem{RatajW18}
Rataj, A., Wozna{-}Szczesniak, B.: Extrapolation of an optimal policy using
  statistical probabilistic model checking. Fundam. Informaticae
  \textbf{157}(4),  443--461 (2018)

\bibitem{shannon1948mathematical}
Shannon, C.E.: A mathematical theory of communication. The Bell system
  technical journal  \textbf{27}(3),  379--423 (1948)

\bibitem{SmolkaKKFHK019}
Smolka, S., Kumar, P., Kahn, D.M., Foster, N., Hsu, J., Kozen, D., Silva, A.:
  Scalable verification of probabilistic networks. In: {PLDI}. pp. 190--203.
  {ACM} (2019)

\bibitem{SB98}
Sutton, R.S., Barto, A.G.: Introduction to Reinforcement Learning. Cambridge,
  MA, USA, 1st edn. (1998)

\bibitem{TapplerA0EL19}
Tappler, M., Aichernig, B.K., Bacci, G., Eichlseder, M., Larsen, K.G.:
  L\({}^{\mbox{*}}\)-based learning of markov decision processes. In: {FM}.
  Lecture Notes in Computer Science, vol. 11800, pp. 651--669. Springer (2019)

\bibitem{Younes02}
Younes, H.L.S., Simmons, R.G.: Probabilistic verification of discrete event
  systems using acceptance sampling. In: {CAV}. Lecture Notes in Computer
  Science, vol.~2404, pp. 223--235. Springer (2002).
  \doi{10.1007/3-540-45657-0\_17},
  \url{https://doi.org/10.1007/3-540-45657-0\_17}

\end{thebibliography}

\newpage
\appendix
\section*{Appendix}
\section{Choice of base Instances}\label{sec:app-base-instances}
\cref{table:baseinstance:choice} shows the effect of different choices of base instances for csma + all\_before\_max. We applied the DT-based policy trained on each such set of  instances and evaluated on the instance with $N=2$ and $K=2$. We see that the base instance set gave the best policy.

\begin{table}[ht]
	\renewcommand{\arraystretch}{1.2}
	\setlength\fboxsep{1cm}
	\caption{Choice of base instances for csma+all\_before\_max. The set used in our experiments is marked as bold.}
	\label{table:baseinstance:choice}
	\begin{adjustbox}{width=\textwidth,center}
		\begin{tabular}{p{0.3\textwidth}|@{\hskip 5mm}p{0.62\textwidth}|@{\hskip 5mm}p{0.2\textwidth}}
			\toprule
			Model instance
			& Parameters for base instances
			& Value \\
			\midrule
			\parbox{0.15\textwidth}{csma+all\_before\_max\\(N=2, K=2)} 
			& \{(N = 2, K = 2)\} & 0.9159110604
			\\
			& \textbf{\{(N = 2, K = 2), (N = 3, K = 2)\}} & \textbf{0.919277227}
			\\
			& \{(N = 2, K = 2), (N = 2, K = 4), (N = 2, K = 6)\} & 0.9159110604
			\\
			& \{(N = 2, K = 2), (N = 2, K = 4)\} & 0.9159110604
			\\
			\bottomrule
			
		\end{tabular}
	\end{adjustbox}
\end{table}

Table~\ref{table:baseinstance} shows the chosen base instances and some information about all the models and properties in our benchmark.

\begin{table}[h]
	\renewcommand{\arraystretch}{1.2}
	\setlength\fboxsep{1cm}
	\caption{MDPs in the benchmark set, parameters, choice of base instances}
	\label{table:baseinstance}
	\begin{adjustbox}{width=\textwidth,center}
		\begin{tabular}{l|@{\hskip 5mm}p{0.6\textwidth} p{0.3\textwidth}}
			\toprule
			Model instance
			& Parameters 
			& \parbox{0.2\textwidth}{Parameter value for base instance} \\
			\midrule
			\multicolumn{3}{c}{Maximizing Probabilities} \\
			consensus+disagree
			& \parbox{0.2\textwidth}{(N, \#modules),\\(K,Var)}
			& \{(N=2,K=8), (N=3,K=8)\} \\
			\hline
			
			csma+all\_before\_max
			& \parbox{0.2\textwidth}{(N, \#modules),\\(K,Var)} 
			& \{(N=2,K=2), (N=3,K=2)\} \\
			\hline
			
			mer+p1
			& \parbox{0.4\textwidth}{(n, Var),\\(x, Var, probability)} 
			& \parbox{0.2\textwidth}{\{(n=1,x=0.01)\}} \\
			\hline 
			
			philosophers+eat
			& (N, \#modules, number of processes)
			& \{(N=4)\}\\
			\hline
			
			zeroconf\_dl+deadline\_max
			& \parbox{0.2\textwidth}{(deadline, Var),\\(K,Var)} 
			& \parbox{0.2\textwidth}{\vspace{3pt}\{(reset=false, \\ N=1000,K=2, \\deadline=50)\}\vspace{3pt}}\\
			\hline
			
			pnueli-zuck+live
			& (N, \#modules, number of processes)
			& \{(N=3)\}\\
			\hline
			
			\multicolumn{3}{c}{Minimizing Probabilities} \\
			consensus+c2
			& \parbox{0.2\textwidth}{(N, \#modules),\\(K,Var)}
			& \{(N=2,K=8), (N=3,K=8)\} \\
			\hline

			csma+some\_before
			& \parbox{0.2\textwidth}{(N, \#modules),\\(K,Var)} 
			& \{(N=2,K=2), (N=3,K=2)\} \\
			\hline
			
			zeroconf\_dl+deadline\_min
			& \parbox{0.2\textwidth}{(deadline, Var),\\(K,Var)} 			
			& \parbox{0.2\textwidth}{\vspace{3pt}\{(reset=false, \\ N=1000,K=2, \\deadline=50)\}\vspace{3pt}} \\
			\hline
			
			firewire+deadline
			& \parbox{0.2\textwidth}{(deadline, Var) , \\ (delay, Var)}
			& \parbox{0.2\textwidth}{\vspace{3pt}\{(deadline=200, \\ delay=3)\}\vspace{3pt}}\\
			\hline
			
			pacman+crash
			& {(MAX\_STEPS, Var)}
			& \{(MAX\_STEPS=5)\} \\
			\bottomrule

		\end{tabular}
	\end{adjustbox}
\end{table}

\section{Parameter Values used in the Experiments}
\label{app:param-values}
\Cref{table:param:values} shows the parameter values we sampled for running our experiments.
We tried sequences based on Fibonacci series, linear, or exponential 
based on the intuition we gained during our initial experiments.

\begin{table}[h]
	\renewcommand{\arraystretch}{1.2}
	\setlength\fboxsep{1cm}
	\caption{The parameter values used in the experiments}
	\label{table:param:values}
	\begin{adjustbox}{width=\textwidth,center}
		\begin{tabular}{l|@{\hskip 5mm}p{0.3\textwidth} p{0.6\textwidth}}
			\toprule
			Model instance
			& Fixed parameters
			& \parbox{0.2\textwidth}{Parameter used for scaling} \\
			\midrule
			\multicolumn{3}{c}{Minimizing Probabilities} \\
			
			zeroconf\_dl+deadline\_min & 	N=1000, K=1 & deadline $\in$ \{100, 200, 300, 600, 1000, 1600, 2000, 2400, 2800, 320\} \\			
			zeroconf\_dl+deadline\_min & 	N=1000, deadline=10 & K $\in$ \{1, 2, 3, 4, 8, 16, 32\}\\			
			firewire+deadline &		deadline=200 & delay $\in$ \{3, 5, 8, 13, 21, 34, 55, 89, 144\}\\			
			firewire+deadline &		delay=3 & deadline $\in$ \{200, 300, 500, 800, 1300, 2100, 3400, 5500, 8900, 14400\}\\
			csma+some\_before &		N=2 & K $\in$ \{2, 3, 5, 6, 7, 8, 10, 11, 12, 13, 17\}\\
			csma+some\_before &		K=2&  N $\in$ \{2, 3, 5, 6, 7, 8, 13, 21, 34, 55, 89, 144\}\\
			consensus+c2&			N=2 & K $\in$ \{2, 3, 5, 8, 13, 21, 34, 55, 89, 144\}\\			
			consensus+c2 & K=2 & N $\in$ \{2, 3, 5, 6, 7, 8, 13, 21, 34, 55, 89, 144\}\\			
			zeroconf\_dl+deadline\_min & deadline=10, K=1 & N $\in$ \{1000, 2000, 4000, 8000, 16000, 32000\}\\	
			pacman+crash & NA & MAX\_STEPs $\in$ \{5, 25, 50, 75, 100, 150, 200, 300, 400, 500\}\\
			
			\midrule
			\multicolumn{3}{c}{Minimizing Probabilities} \\
			
			zeroconf\_dl+deadline\_max 	&	N=1000, K=1			& deadline $\in$ \{100, 200, 300, 600, 1000, 1600, 2000, 2400, 2800, 320\} \\
			zeroconf\_dl+deadline\_max 	&	N=1000, deadline=10			& K $\in$ \{1, 2, 3, 4, 8, 16, 32\}\\
			zeroconf\_dl+deadline\_max  	&	deadline=10, K=1& N $\in$ \{1000, 2000, 4000, 8000, 16000, 32000\}\\
			philosophers+eat		&	& N $\in$ \{3, 4, 5, 8, 13, 21, 24, 27, 30, 34, 55, 89, 144\} \\
			pnueli-zuck+live		&	& N $\in$ \{3, 4, 5, 8, 13, 21, 24, 27, 30, 34, 55, 89, 144\} \\
			csma+all\_before\_max 		&	 N=2			& K $\in$ \{2, 3, 5, 6, 7, 8, 10, 11, 12, 13, 17\}\\
			mer+p1 				&	 x=0.01& n $\in$ \{1, 10, 100, 1000, 3000, 8000, 13000, 21000, 34000, 55000, 89000, 144000\}\\
			mer+p1 				& n=10& x $\in$  \{0.01, 0.03, 0.05, 0.08, 0.13, 0.21, 0.34, 0.55, 0.89\}\\
			consensus+disagree 		& K=2& N $\in$ \{2, 3, 5, 6, 7, 8, 13, 21, 34, 55, 89, 144\}\\			
			consensus+disagree 		&	N=2& K $\in$ \{2, 3, 5, 8, 13, 21, 34, 55, 89, 144\}\\
			csma+all\_before\_max 		& K=2 &  N $\in$ \{2, 3, 5, 6, 7, 8, 13, 21, 34, 55, 89, 144\}\\
			
		\end{tabular}
	\end{adjustbox}
\end{table}

\section{Result of simulations using random schedulers}
Table~\ref{table:results:min:random} and Table~\ref{table:results:max:random} reports out results from running $1000$ simulations using randomly generated $1000$ independent schedulers. 
A model and property combination with higher coefficient of variation (ratio of the standard deviation to the mean) would imply that the distribution of policies are more sparse
and it is improbable to find a near optimal policy by randomly generating one. 

\begin{table}[t!]
	\renewcommand{\arraystretch}{1.2}
	\setlength\fboxsep{1cm}
	\caption{The results table for minimization properties. This is generated by simulating $1000$ randomly generated independent schedulers. MaxV and MinV refers to maximum and minimum over the empirical value calculated for these schedulers by simulating $1000$ simulations. \oor\ means out-of-resources (both time and memory) and \rle\ means \emph{run length exceeded}, 
	an exception given by \modes
	}
	\label{table:results:min:random}
	\begin{adjustbox}{width=1.2\textwidth,center}	\begin{tabular}{p{0.3\textwidth}|lc|S[table-format=-1.2e-2,table-align-text-post=false] S[table-format=-1.2e-2,table-align-text-post=false] S[table-format=-1.2e-2,table-align-text-post=false] S[table-format=-1.2e-2,table-align-text-post=false] S[table-format=-1.2e-2,table-align-text-post=false] }
			\toprule
			
			& \multicolumn{2}{c|}{\textbf{Scale}} &  \multicolumn{5}{c}{\textbf{Results of 1000 schedulers, 1000 runs}}\\
			
			Model+propery 
			& Variable
			& Time
			& {Mean $(\mu)$}
			& {Variance $(\sigma^2)$}
			& {MaxV}
			& {MinV}
			& {Coeff. Var $(\frac{\sigma}{\mu})$} \\\midrule
			
			\multirow{3}{*}{\parbox{0.3\textwidth}{zeroconf\_dl+deadline\_min\\
					(N=1000, K=1)}}
			& dl=200 &\aminute  &  0.0006670000000000005 & 			\expnum{8.16111000000006e-07} & 			0.006 & 0 \notemodest& 	0.0012235547226386889 \\
			& dl=1600&\anhour  &  0.0006550000000000004 &\expnum{9.059749999999887e-07} &0.007 & 0 \notemodest& 0.0013831679389312795\\
			& dl=3200&Beyond  & 0.0007180000000000005	&\expnum{9.304760000000008e-07} 	&0.005 &0 \notemodest&0.0012959275766016715\\
			\midrule			
			
			\multirow{3}{*}{\parbox{0.3\textwidth}{zeroconf\_dl+deadline\_min\\
					(N=1000, deadline=10)} }
			& K=2 &\aminute & 0.3422200000000006 &0.0002187295999999994 &0.386 &0.297 \notemodest	&0.0006391490853836684\\
			& K=8  & \aminute &   1  &0 &1 &1 &0 \\[5pt]
			\midrule
			
			\multirow{3}{*}{\parbox{0.3\textwidth}{firewire+deadline\\ (deadline=200)}}
			& delay=5& \aminute & 0.9909990000000001
			&0.0020474429990000225
			&1.0
			&0.718
			&0.0020660394198178025\\
			&delay=21 &\aminute  &   1  &0 &1 &1 &0\\
			&delay=34 &\aminute  &  1  &0 &1 &1 &0\\
			&delay=89 &\aminute  &   1  &0 &1 &1 &0\\			
			\midrule
			
			\multirow{3}{*}{\parbox{0.3\textwidth}{firewire+deadline\\ (delay=3)}}
			& dl=200  &  \aminute & 0.9709269999999999
			&0.00629393967100013
			&1.0
			&0.547
			&0.006482402560645786 \\
			& dl=500 & \aminute  &   1  &0 &1 &1 &0\\			
			& dl=1300 & \aminute &   1  &0 &1 &1 &0\\			
			\midrule
			
			\multirow{3}{*}{\parbox{0.3\textwidth}{csma+some\_before\\ (N=2)}}
			& K=2 & \aminute  &  0.5003709999999995
			&0.00025218535900000023
			&0.549
			&0.451\notemodest
			& 0.000503996752409713 \\
			& K=3 & \aminute  & 0.8747569999999995
			&0.00011404195100000005
			&0.907
			&0.84
			&0.00013036986385933479 \\
			\midrule
			
			\multirow{3}{*}{\parbox{0.3\textwidth}{csma+some\_before\\ (fix K=2)}}
			& N=3 & \aminute  & 0.9025649999999997 & 0.01902298377500026 & 1.0 & 0.56\notemodest & 0.021076580384792525\\
			& N=5 & \anhour  & 0.7778539999999998 & 0.031070636684000143 & 1.0 & 0.452 & 0.03994404693425778\\
			& N=8 & \anhour  &  0.7602989999999993 & 0.022226469599000095 & 1.0 & 0.519 & 0.029233853522101325 \\
			& N=13 & Beyond  &  0.7618870000000004 & 0.01168564423099999 & 1.0 & 0.519 & 0.015337765614848375 \\
			\midrule
			
			\multirow{3}{*}{\parbox{0.3\textwidth}{consensus+c2\\ (N=2)}}
			& K=2 & \aminute &  0.48594900000000113 & 0.001179066398999999 & 0.594 & 0.381\notemodest & 0.00242631716291215\\
			& K=55  & \aminute &  \error  &  \error & \error  & \error & \error \\
			& K=144  & \aminute & \error  &  \error  & \error  & \error & \error \\
			\midrule
			
			\multirow{3}{*}{\parbox{0.3\textwidth}{consensus+c2\\ (K=2)}}
			& N=6 &\aminute  &  0.4831310000000007 & 0.00026700783900000097 & 0.537 & 0.43 & 0.000552661367206825 \\
			& N=7 &\anhour  & 0.48219200000000145 & 0.0002398211359999999 & 0.538 & 0.436 & 0.0004973561071108587 \\
			& N=13 & Beyond  &  \error  &  \error  &  \error & 0.5  &  \error \\
			\midrule

			\multirow{3}{*}{\parbox{0.3\textwidth}{zeroconf\_dl+deadline\_min\\
					(deadline=10, K=1)} }
			& N=1000 &\aminute &  0.006983000000000012 & \expnum{1.0036710999999929e-05} & 0.019 & 0\notemodest & 0.0014373064585421612\\
			& N=8000  & \aminute & 0.060227000000000086 & 0.00020765747099999913 & 0.126 & 0.022 & 0.0034479132448901463\\
			& N=32000 & \aminute& 0.31968900000000067 & 0.001492770278999998 & 0.467 & 0.209 & 0.004669445238966605\\
			\midrule
			
			\multirow{3}{*}{\parbox{0.3\textwidth}{pacman+crash$\downarrow$}}
			& MS=5 &\aminute &  0.5507580000000006 & 0.0002433894359999982 & 0.605 & 0.508\notemodest & 0.0004419172050156292 \\
			& MS=25  & \aminute & 0.9226340000000002 & 0.0014816540439999996 & 1.0 & 0.693 & 0.0016058957766568317 \\
			& MS=200 & \anhour &  1  &0 &1 &1 &0  \\
			& MS=300 & Beyond &  1  &0 &1 &1 &0\\
			
			\bottomrule
		\end{tabular}
	\end{adjustbox}
\end{table}

\begin{table}[t!]
	\renewcommand{\arraystretch}{1.2}
	\setlength\fboxsep{1cm}
	\caption{The results table for maximization properties. This is generated by simulating $1000$ randomly generated independent schedulers. MaxV and MinV refers to maximum and minimum over the empirical value calculated for these schedulers by simulating $1000$ simulations. \oor\ means out-of-resources (both time and memory) and \rle\ means \emph{run length exceeded}, an exception that occurs 
		when the scheduler gets trapped in an end component.}
	\label{table:results:max:random}
	\begin{adjustbox}{width=1.2\textwidth,center}	\begin{tabular}{p{0.3\textwidth}|lc|S[table-format=-1.2e-2,table-align-text-post=false] S[table-format=-1.2e-2,table-align-text-post=false] S[table-format=-1.2e-2,table-align-text-post=false] S[table-format=-1.2e-2,table-align-text-post=false] S[table-format=-1.2e-2,table-align-text-post=false] }
			\toprule

			& \multicolumn{2}{c|}{\textbf{Scale}} &  \multicolumn{5}{c}{\textbf{Results of 1000 schedulers, 1000 runs}}\\
			
			Model+propery 
			& Variable
			& Time
			& {Mean $(\mu)$}
			& {Variance $(\sigma^2)$}
			& {MaxV}
			& {MinV}
			& {Coeff. Var $(\frac{\sigma}{\mu})$} \\\midrule

			\multirow{3}{*}{\parbox{0.3\textwidth}{zeroconf\_dl+deadline\_max  \\	(N=1000, K=1)}}
			&	dl=200& \aminute  & 0.0007530000000000005 & \expnum{9.619909999999998e-07} & 0.006\notemodest & 0.0 & 0.0012775444887118182\\
			& dl=1600& \anhour  &  0.0006550000000000004 & \expnum{8.619749999999887e-07} & 0.007 \notemodest& 0.0 & 0.0013159923664121957\\
			& dl=3200& Beyond  &  0.0006360000000000005 & \expnum{8.315039999999931e-07} & 0.005 & 0.0 & 0.0013073962264150826 \\
			\midrule
			
			\multirow{3}{*}{\parbox{0.3\textwidth}{zeroconf\_dl+deadline\_max \\
					(N=1000, deadline=10)} }
			& K=2  & \aminute &  0.3418300000000005 & 0.0002463031000000002 & 0.385 \notemodest& 0.299 & 0.0007205426674077753\\
			& K=8  & \aminute &  1  &0 &1 &1 &0\\
			& K=32  & \aminute &  1  &0 &1 &1 &0  \\
			\midrule
			
			\multirow{3}{*}{\parbox{0.3\textwidth}{zeroconf\_dl+deadline\_max  \\
					(deadline=10, K=1)} }
			& N=1000  & \aminute & 0.006871999999999993 & \expnum{9.585615999999989e-06} & 0.019\notemodest & 0 & 0.001394880093131548\\
			& N=8000   & \aminute & 0.06108600000000007 & 0.00019169860400000002 & 0.117 & 0.022 & 0.003138175752218181\\
			& N=32000 & \aminute & 0.3197710000000013 & 0.0014185245589999993 & 0.451 & 0.209 & 0.004436063805035458 \\
			\midrule
			
			\multirow{3}{*}{\parbox{0.3\textwidth}{philosophers+eat }}
			&N=5 & \aminute  &  0.03734400000000001 & 0.01894231966399999 & 0.934 & 0.0 & 0.5072386371036842\\
			&N=21& \anhour  &  0.000245 & \expnum{5.9964975e-05} & 0.245 & 0.0 & 0.244755\\
			&N=34& Beyond  &   0 &   0 &  0 & 0 & 0 \\
			\midrule
			
			\multirow{3}{*}{\parbox{0.3\textwidth}{pnueli-zuck+live }}
			&N=5 &\aminute  &  0.035749 & 0.034028273998999964 & 1.0 & 0.0 & 0.9518664577750415\\
			&N=21 &\anhour  & 0.002 & 0.001996 & 1 & 0 & 0.998\\
			&N=34 & Beyond  &  0.002 & 0.001996 & 1 & 0 & 0.998\\
			\midrule			
			
			\multirow{3}{*}{\parbox{0.3\textwidth}{csma+all\_before\_max  \\ (N=2)}}
			& K=8 & \aminute  &   1  &0 &1 &1 &0\\
			& K=11 & \anhour  &    1  &0 &1 &1 &0\\
			& K=13 & Beyond  &  \oor  &  \oor &  \oor & \oor & \oor \\
			\midrule

			\multirow{3}{*}{\parbox{0.3\textwidth}{mer+p1  \\ (x=0.01)}}
			& n=1000& \aminute  &  0.000184 & \expnum{3.3822144e-05} & 0.184 & 0.0 & 0.183816\\
			& n=21000 &\anhour  &  0 & 0 & 0 & 0 & 0\\
			& n=55000 & Beyond  &  0.000399 & \expnum{7.9441799e-05} & 0.2 & 0.0 & 0.19910225313283209\\
			\midrule

			\multirow{3}{*}{\parbox{0.3\textwidth}{mer+p1  \\ (n=10)}}
			& x=0.01  &  \aminute & 0.000402 & \expnum{8.1890396e-05} & 0.226\notemodest & 0 & 0.20370745273631838\\
			& x=0.08  & \aminute & 0.000175 & \expnum{3.059437499999999e-05} & 0.175 & 0 & 0.17482499999999995\\
			& x=0.55  &  \aminute & 0.000494 & \expnum{6.7093964e-05} & 0.191 & 0 & 0.13581774089068827 \\
			& x=0.89 & \aminute  & 0.0005809999999999999 & 0.00011244143900000001 & 0.203 & 0 & 0.19353087607573155 \\
			\midrule

			\multirow{3}{*}{\parbox{0.3\textwidth}{consensus+disagree  \\ (K=2)}}
			& N=5 & \aminute  &  0.03502900000000011 & \expnum{5.2366158999999534e-05} & 0.063 & 0.007 & 0.0014949373090867388 \\
			& N=7 &\anhour  &  0.03409800000000015 & \expnum{3.530239599999999e-05} & 0.053 & 0.018 & 0.0010353216024400211\\
			& N=13& Beyond  &  \rle  &  \rle  &  \rle & \rle& \rle \\
			\midrule			
			
			\multirow{3}{*}{\parbox{0.3\textwidth}{consensus+disagree  \\ (N=2)}}
			& K=2  & \aminute& 0.027651000000000005 & 0.0005231511990000013 & 0.104 & 0 & 0.018919793099707104\\
			& K=55  &  \aminute &\rle & \rle & \rle & \rle & \rle \\
			& K=144  &  \aminute& \rle  &  \rle  &  \rle & \rle & \rle \\
			\midrule
			
			\multirow{3}{*}{\parbox{0.3\textwidth}{csma+all\_before\_max  \\ (K=2)}}
			& N=3 &\aminute  &  0.6838249999999996 & 0.0061983063750000135 & 0.865\notemodest & 0.512 & 0.00906417047490223\\
			& N=5 &\anhour  & 0.2363679999999993 & 0.001973376576000002 & 0.43 & 0.143 & 0.008348746767752012\\
			& N=6& Beyond  & 0.11185200000000006 & 0.0007153820959999957 & 0.216 & 0.059 & 0.0063957917247791304 \\
			
			\bottomrule
		\end{tabular}
	\end{adjustbox}
\end{table}

\section{Time comparison}
\label{app:time}
Table~\ref{table:results:min:time} and Table~\ref{table:results:max:time} shows the time needed for creating and evaluating the DT on different instances.
\sisetup{round-mode=places,round-precision=2}
\begin{table}[t!]
	\caption{
		The time for constructing and evaluating the decision tree, and computing the stategy using \modest smart LSS for \emph{minimizing} properties. 
		We shorten the parameters delay to \emph{d}, deadline to \emph{dl}, and MAX\_STEPS to \emph{MS}.
		\oor\ means out-of-resources (both time and memory) and \rle\ means \emph{run length exceeded}, 
		an exception given by \modes.
	}
	\label{table:results:min:time}
	\begin{adjustbox}{width=1\textwidth,center}
		\begin{tabular}{p{0.3\textwidth}|lc|
				r
				r ||
				c
			}
			\toprule
			
			Model+property & \multicolumn{2}{c|}{\textbf{Scale}} & \multicolumn{3}{c}{\textbf{Time (s)}}\\
			
			(values~of~parameters) 
			& Variable
			& Time
			& {\shortstack{DT \\ Synthesis}}
			& {\shortstack{Smart \\ LSS}}
			& {\shortstack{DT precise \\ evaluation}}
			\\\midrule
			
			\multirow{3}{*}{\parbox{0.3\textwidth}{zeroconf\_dl+deadline\_min\\
					(N=1000, K=1)}}
			& dl=200 &\aminute  &  184.823  & 3.52\notemodest  & 8.42    \\
			& dl=1600&\anhour  &  184.823   & 3.48  &  61.7  \\
			& dl=3200&Beyond  &  184.823  &  3.53 &  134.0   \\
			\midrule			
			
			\multirow{3}{*}{\parbox{0.3\textwidth}{zeroconf\_dl+deadline\_min\\
					(N=1000, deadline=10)} }
			& K=2 &\aminute &  184.823 & 2.89 & 2.04 \\
			& K=8  & \aminute & 184.823 & 4.09 & 2.04 \\[5pt]
			
			
			\midrule
			
			\multirow{3}{*}{\parbox{0.3\textwidth}{firewire+deadline\\ (deadline=200)}}
			& delay=5& \aminute &  0.657 & 11.1 & .743 \\
			&delay=21 &\aminute  & 0.657     & 9.15  &  .629 \\
			&delay=34 &\aminute  & 0.657     & 8.1  &  .644 \\
			&delay=89 &\aminute  &  0.657    & 5.42  &  .523 \\			
			\midrule
			
			\multirow{3}{*}{\parbox{0.3\textwidth}{firewire+deadline\\ (delay=3)}}
			& dl=200  &  \aminute & 0.657   & 11.9\notemodest  &  .742  \\
			& dl=500 & \aminute & 0.657  & 10.2 & 1.85\\
			& dl=1300 & \aminute& 0.657 & 9.94 & 9.42 \\
			\midrule
			
			\multirow{3}{*}{\parbox{0.3\textwidth}{csma+some\_before\\ (N=2)}}
			& K=2 & \aminute  & 3.437    & 8.32\notemodest  & .645 \\
			& K=3 & \aminute  & 3.437    & 7.13  &  1.26 \\
			\midrule
			
			\multirow{3}{*}{\parbox{0.3\textwidth}{csma+some\_before\\ (fix K=2)}}
			& N=3 & \aminute  & 3.437    & 15.6\notemodest &  4.12 \\
			& N=5 & \anhour  &  3.437    & 54.8  &  957.00 \\
			& N=8 & \anhour  &  3.437    & 167.00  &  20392.09 {$^\dagger$} \\
			& N=13 & Beyond  &  3.437  &  620.00 &  \oor  \\
			\midrule
			
			\multirow{3}{*}{\parbox{0.3\textwidth}{consensus+c2\\ (N=2)}}
			& K=2 & \aminute & 4.067    & 5.36 &  .348 \\
			& K=55  & \aminute &  4.067 & \rle  &  .812 \\
			& K=144  & \aminute & 4.067  & \rle  &  1.65  \\
			\midrule
			
			\multirow{3}{*}{\parbox{0.3\textwidth}{consensus+c2\\ (K=2)}}
			& N=6 &\aminute  & 4.067   & 89.0  &  2.06 \\
			& N=7 &\anhour  &  4.067 & 141.0  &  5.76  \\
			& N=13 & Beyond  &  4.067    &  \rle &  467.0 \\
			\midrule
			
			\multirow{3}{*}{\parbox{0.3\textwidth}{zeroconf\_dl+deadline\_min\\
					(deadline=10, K=1)} }
			& N=1000 &\aminute & 184.823 & 3.57 & 1.68 \\
			& N=8000  & \aminute & 184.823 & 1.68  & 1.65\\
			& N=32000 & \aminute& 184.823 & 3.12 & 1.63   \\
			\midrule

			\multirow{3}{*}{\parbox{0.3\textwidth}{pacman+crash}}
			& MS=5 &\aminute &  0.29 & 7.65\notemodest & 2.55 \\
			& MS=25  & \aminute & 0.29 & 18.9 & 4.49  \\
			& MS=200 & \anhour & 0.29 & 16.8 & 247.0  \\
			& MS=300 & Beyond & 0.29 & 17.1& 396.0 \\
			
			\bottomrule
		\end{tabular}
	\end{adjustbox}
\end{table}

\sisetup{round-mode=places,round-precision=2}
\begin{table}[t!]
	\setlength\fboxsep{1cm}
	\caption{
		The time for constructing and evaluating the decision tree, and computing the stategy using \modest smart LSS for \emph{maximizing} properties.
\rle \ means \emph{run length exceeded} given by \modes.
	}
	\label{table:results:max:time}
	\begin{adjustbox}{width=1\textwidth,center}
	\begin{tabular}{p{0.3\textwidth}|lc|
			r
			r ||
			r
			HH
		}
		\toprule
		
		Model+property & \multicolumn{2}{c|}{\textbf{Scale}} & \multicolumn{3}{c}{\textbf{Time (s)}}\\
		
		(values~of~parameters) 
		& Variable
		& Time
		& {\shortstack{DT \\ Synthesis}}
		& {\shortstack{Smart \\ LSS}}
		& {\shortstack{DT precise \\ evaluation}}
		\\\midrule

			\multirow{3}{*}{\parbox{0.3\textwidth}{zeroconf\_dl+deadline\_max \\	(N=1000, K=1)}}
			&	dl=200& \aminute  & 205.719    & 1.55  &  14.2 \\
			& dl=1600& \anhour  &  205.719    & 1.54  &  72.9 \\
			& dl=3200& Beyond  &  205.719    & 1.55  &  145.0 \\
			\midrule
			
			\multirow{3}{*}{\parbox{0.3\textwidth}{zeroconf\_dl+deadline\_max \\
					(N=1000, deadline=10)} }
			& K=2  & \aminute &  205.719 & 3.35 & 4.01 \\
			& K=8  & \aminute &  205.719  & 4.10 & 4.62 \\
			& K=32  & \aminute &  205.719  & 4.23 & 4.64  \\
			\midrule
			
			\multirow{3}{*}{\parbox{0.3\textwidth}{zeroconf\_dl+deadline\_max  \\
					(deadline=10, K=1)} }
			& N=1000  & \aminute & 205.719  & 1.88 & 4.01\\
			& N=8000   & \aminute & 205.719  & 2.42 & 4.00\\
			& N=32000 & \aminute & 205.719  & 3.64 & 4.00\\
			\midrule
			
			\multirow{3}{*}{\parbox{0.3\textwidth}{philosophers+eat}}
			&N=5 & \aminute  &  1.307   & 3.03  &  .447  \\
			&N=21& \anhour  &   1.307   & 12.7  &  1.35 \\
			&N=34& Beyond  &   1.307  &  22.4 &   2.07 \\
			\midrule
			
			\multirow{3}{*}{\parbox{0.3\textwidth}{pnueli-zuck+live}}
			&N=5 &\aminute  &  0.415     & 4.24  & .785 \\
			&N=21 &\anhour  &  0.415     & 37.4  & 7.99 \\
			&N=34 & Beyond  &  0.415    & 128.0  & 19.6  \\
			\midrule			
			
			\multirow{3}{*}{\parbox{0.3\textwidth}{csma+all\_before\_max \\ (N=2)}}
			& K=8 & \aminute  &  3.684   & 35.9  & 911.0  \\
			& K=11 & \anhour  &  3.684    & 171.0  &  321.0 {$^\dagger$}\\
			& K=13 & Beyond  &  3.684   &  666.0 &  1200.0 {$^\dagger$} \\
			\midrule

			\multirow{3}{*}{\parbox{0.3\textwidth}{mer+p1 \\ (x=0.01)}}
			& n=1000& \aminute  &  1.027    & \rle  & 0.375\\
			& n=21000 &\anhour  &  1.027  &  0.410  & \rle  &  0.410\\
			& n=55000 & Beyond  &  1.027   & \rle  & 0.375 \\
			\midrule
			
			\multirow{3}{*}{\parbox{0.3\textwidth}{mer+p1 \\ (n=10)}}
			& x=0.01  &  \aminute & 1.027    & \rle  & 0.409\\
			& x=0.08  & \aminute & 1.027     & \rle  &  0.404\\
			& x=0.55  &  \aminute & 1.027   & \rle  & 0.387 \\
			& x=0.89 & \aminute  &  1.027    & \rle  & 0.416\\			
			\midrule

			\multirow{3}{*}{\parbox{0.3\textwidth}{consensus+disagree \\ (K=2)}}
			& N=5 & \aminute  &  0.392   &  36.1 &  1.71 \\
			& N=7 &\anhour  &  0.392    &  108.0 &  5.28 \\
			& N=13& Beyond  &  0.392    &  \rle &  0.670 \\
			\midrule			
			
			\multirow{3}{*}{\parbox{0.3\textwidth}{consensus+disagree \\ (N=2)}}
			& K=2  & \aminute& 0.392   & 3.54 &  0.250\\
			& K=55  &  \aminute & 0.392 & \rle & 1.17  \\
			& K=144  &  \aminute& 0.392    &  \rle &  3.58 \\
			\midrule
			
			\multirow{3}{*}{\parbox{0.3\textwidth}{csma+all\_before\_max \\ (K=2)}}
			& N=3 &\aminute  &  3.684   & 16.8  &  4.60  \\
			& N=5 &\anhour  &  3.684    &  48.9 &  1220.0 \\
			& N=6& Beyond  &  3.684    & 49.7  & 1690.0{$^\dagger$} \\
			
			\bottomrule
		\end{tabular}
	\end{adjustbox}
\end{table}

\end{document}